\documentclass{article}

\usepackage{microtype}
\usepackage{graphicx}
\usepackage{booktabs} % for professional tables

\usepackage{hyperref}

\usepackage{amsmath}
\usepackage{amssymb}
\usepackage{mathtools}
\usepackage{amsthm}

\usepackage{xcolor}  
\usepackage{colortbl}
\definecolor{linkcolor}{RGB}{83,83,182}
\hypersetup{
    colorlinks=true,
    citecolor=linkcolor,
    linkcolor=linkcolor
}
\usepackage{arydshln} % Add this to use \hdashline

\newcommand{\ft}{\algname{$R^2$-DSnoT}}

\DeclareMathOperator*{\argmax}{arg\,max}

\usepackage{graphicx} % For including graphics

\usepackage{caption}
\usepackage{subcaption}
   
\usepackage{xcolor}
\usepackage{mathtools}
\newcommand{\eqdef}{\coloneqq}
\usepackage{pifont}
\definecolor{mygreen1}{rgb}{0,0.8,0}
\newcommand{\green}{\color{mydarkgreen}}
\newcommand{\red}{\color{mydarkred}}
\usepackage{algorithm}
\definecolor{bgcolor}{rgb}{0.93,0.99,1}
\definecolor{bgcolor2}{rgb}{0.8,1,0.8}
\definecolor{bgcolor3}{rgb}{0.50,0.90,0.50}
\definecolor{bgcolor4}{rgb}{0.94,0.97,1}

\usepackage{xspace}
\usepackage{scalefnt}
\DeclareMathOperator*{\minimize}{\mathrm{minimize}}
\newcommand{\ec}[1]{\mathbb{E}\!\left[ #1 \right]}

\definecolor{Gray}{gray}{0.95}

\definecolor{mydarkgreen}{rgb}{0,0.42,0}
\definecolor{mydarkred}{rgb}{0.75,0,0}
\definecolor{mygreen2}{RGB}{0,120,20}
\newcommand{\algname}[1]{{\sf\scalefont{0.90}{#1}}\xspace}
\usepackage{booktabs}

\newcommand{\sqn}[1]{{\left\lVert#1\right\rVert}^2}
\newcommand{\norm}[1]{\left\| #1 \right\|}
\newcommand{\mbR}{\mathbb{R}}

\usepackage{pifont}
\newcommand{\cmark}{\green\ding{51}}%
\newcommand{\xmark}{\red\ding{55}}%

\newcommand{\vA}{{\mathbf{A}}}
\newcommand{\vB}{{\mathbf{B}}}
\newcommand{\vC}{{\mathbf{C}}}
\newcommand{\vD}{{\mathbf{D}}}

\newcommand{\vG}{{\mathbf{G}}}

\newcommand{\vI}{{\mathbf{I}}}

\newcommand{\vS}{{\mathbf{S}}}

\newcommand{\vW}{{\mathbf{W}}}
\newcommand{\vX}{{\mathbf{X}}}
\newcommand{\vY}{{\mathbf{Y}}}

\usepackage[normalem]{ulem}

\newcommand{\Diag}{\mathrm{Diag}}

\usepackage{multirow}
\usepackage{threeparttable}
\usepackage{nicefrac}
      
\usepackage[accepted]{icml2024}

\usepackage[capitalize,noabbrev]{cleveref}

\theoremstyle{plain}
\newtheorem{theorem}{Theorem}[section]

\newtheorem{lemma}[theorem]{Lemma}
\newtheorem{corollary}[theorem]{Corollary}
\theoremstyle{definition}

\theoremstyle{remark}

\usepackage[textsize=tiny]{todonotes}

\icmltitlerunning{Symmetric Pruning of Large Language Models}

\begin{document}

\twocolumn[
% \icmltitle{Symmetric Post-Training Pruning and Training-Free Fine-Tuning for Large Language Models}
% \icmltitle{Symmetric Pruning and Training-Free Fine-Tuning for Large Language Models}
% \icmltitle{Symmetric Post-Training Pruning for Large Language Models}
\icmltitle{Symmetric Pruning for Large Language Models}
% \icmltitle{Theory-Guided Symmetric Pruning for Large Language Models}




\begin{icmlauthorlist}
\icmlauthor{Kai Yi}{yyy}
\icmlauthor{Peter Richtárik}{yyy}
\end{icmlauthorlist}

\icmlaffiliation{yyy}{Department of Computer Science, King Abdullah University of Science and Technology (KAUST), Thuwal, Saudi Arabia.}


\icmlcorrespondingauthor{Kai Yi}{kai.yi@kaust.edu.sa}

\icmlkeywords{Machine Learning, ICML}

\vskip 0.3in
]



\printAffiliationsAndNotice{}  % leave blank if no need to mention equal contribution

\begin{abstract} 
Popular post-training pruning methods such as \algname{Wanda} \citep{Wanda} and \algname{RIA} \citep{RIA} are known for their simple, yet effective, designs that have shown exceptional empirical performance. \algname{Wanda} optimizes performance through calibrated activations during pruning, while \algname{RIA} emphasizes the relative, rather than absolute, importance of weight elements. Despite their practical success, a thorough theoretical foundation explaining these outcomes has been lacking. This paper introduces new theoretical insights that redefine the standard minimization objective for pruning, offering a deeper understanding of the factors contributing to their success. Our study extends beyond these insights by proposing complementary strategies that consider both input activations and weight significance. We validate these approaches through rigorous experiments, demonstrating substantial enhancements over existing methods. Furthermore, we introduce a novel training-free fine-tuning approach \ft that incorporates relative weight importance and a regularized decision boundary within a dynamic pruning-and-growing framework, significantly outperforming strong baselines and establishing a new state-of-the-art. \end{abstract}
% This work not only bridges the gap between theory and practice but also paves the way for more robust pruning methodologies in large language models. 

\section{Introduction} Large Language Models (LLMs) \citep{OPT, LlaMA, LlaMA2, Phi2} have demonstrated remarkable capabilities across a variety of tasks. However, their extensive size often hinders practical deployment. Interest in LLM compression has surged in recent years, driven by the need to reduce model sizes while maintaining performance \citep{SmoothQuant, SparseGPT, Wanda, RIA, PV-Tuning}. This paper focuses on LLM \textbf{post-training pruning (PTP)}, a prevalent method for reducing the footprint of pre-trained weights.

%Consider a target sparsity ratio $\epsilon$, a pre-trained model weight $\vW$, and a pruned weight $\widetilde{\vW}$. Our goal is to minimize an optimization objective $F(\widetilde{\vW} | \vW, \epsilon)$, subject to the constraint $\operatorname{Mem}(\widetilde{\vW}) \leq \epsilon \operatorname{Mem}(\vW)$. 
A common approach to pruning is magnitude-based pruning, where elements of each layer's weights with smaller absolute values are set to zero. In contrast, \algname{Wanda} \citep{Wanda} introduced an innovative method that scales the weights by the activations of each layer, demonstrating promising performance on standard benchmarks. Building upon this, \algname{RIA} \citep{RIA} further improved the approach by evaluating the relative importance of each weight across its corresponding row and column before pruning. While their empirical results are encouraging, the underlying mechanisms remain poorly understood. This leads us to our first question:

\emph{Can we provide theoretical support for post-training pruning methods and derive more efficient algorithms with minimal adaptations to the existing framework?}

To deepen our understanding of these popular PTP methods, we introduce a novel formulation—referred to as \textbf{Sym}metric \textbf{W}eight \textbf{And} \textbf{A}ctivation (\algname{SymWanda})—that aims to efficiently leverage \textit{both} the input activation of a layer and the output for that layer. This symmetric and generalized approach provides theoretical insights into the mechanisms of established empirical methods such as \algname{Wanda} and \algname{RIA}. 

Intrinsic PTP methods have demonstrated remarkable performance, as reflected by perplexity scores and zero-shot accuracy. However, their performance can degrade significantly when the sparsity ratio is high. This is due to the intrinsic reconstruction error between the pruned weights and the original pre-trained weights. Minimizing this reconstruction error is particularly important for efficient post-training pruning. Beyond LLM pruning, we explore further fine-tuning to enhance model efficiency and performance. This brings us to our second problem:

\emph{Can we fine-tune pruned LLMs without further training and outperforms state-of-the-art methods with minimal effort?}

\textbf{Dynamic sparse training (DST)} has gained attention for selectively updating and maintaining a subset of network parameters throughout the training process while dynamically adapting the sparse topology through weight operations. Its proven efficiency in enabling effective training suggests DST could be a promising approach for fine-tuning LLMs in an efficient manner. However, DST inherently requires backpropagation to train subnetworks, and its effectiveness heavily depends on a sufficient number of weight updates \citep{liu2021we}. %Moreover, studies have reported its limitations when applied to fine-tuning small-scale BERT-level models \citep{liu2023sparsity}.

Interestingly, the pruning-and-growing step within DST offers a training-free methodology, where sparse mask adaptation is based solely on weight properties such as magnitude \citep{mocanu2018scalable}. This opens up a potential alternative for addressing the challenge: Instead of relying on computationally intensive backpropagation for fine-tuning sparse LLMs, we can explore the iterative updating of sparse masks in a training-free manner. Motivated by this insight, we focus on training-free fine-tuning approaches.

\algname{DSnoT} \citep{zhang2023dynamic} introduced a straightforward yet effective method for pruning and growing weights using their values and statistical metrics (e.g., expectation and variance) for each ongoing pruning row. Inspired by \algname{Wanda}, \algname{DSnoT} achieves simplicity but falls short of fully leveraging relative weight information, particularly in scenarios where weight distributions are highly non-uniform and contain many outliers \citep{RIA}. To address these limitations, we propose incorporating relative weight importance into the growing criterion design. Furthermore, we observe that directly optimizing for reconstruction error is suboptimal. To improve performance, we introduce a regularization term that relaxes the decision boundary.
Our new designs demonstrate significant efficiency and consistently achieve promising performance, paving the way for more effective and computationally feasible fine-tuning methods for sparse LLMs.

% Our \textbf{contributions} are summarized as follows:

% $\bullet$ We propose a novel formulation, \algname{SymWanda}, which minimizes the impact of pruning on both input activations and output influences of weights. This approach provides theoretical insights into the empirical successes of methods such as \algname{Wanda} and \algname{RIA}.

% $\bullet$Building on this formulation, we introduce a series of innovative pruning strategies. Extensive experiments validate the effectiveness of our methods. Notably, we incorporate an efficient stochastic approach for manipulating relative importance, which achieves superior performance with highly reduced sampling cost.

% $\bullet$ We present a novel training-free fine-tuning method \ft that leverages relative weight importance and a regularized decision boundary within a pruning-and-growing framework. This approach significantly outperforms strong baselines, achieving remarkable results.

Our \textbf{contributions} are summarized as follows: 
\textbf{i):} We propose a novel formulation, \algname{SymWanda}, which minimizes the impact of pruning on both input activations and output influences of weights. This approach provides theoretical insights into the empirical successes of methods such as \algname{Wanda} and \algname{RIA}. 
\textbf{ii):} Building on this formulation, we introduce a series of innovative pruning strategies. Extensive experiments validate the effectiveness of our methods. Notably, we incorporate an efficient stochastic approach for manipulating relative importance, which achieves superior performance with highly reduced sampling cost. 
\textbf{iii):} We present a novel training-free fine-tuning method \ft that leverages relative weight importance and a regularized decision boundary within a pruning-and-growing framework. This approach significantly outperforms strong baselines, achieving remarkable results.

% \end{itemize}

% \section{Related Work}
 
\begin{table*}[t]
    \centering
    \scriptsize
    \caption{\footnotesize Comparison of LLM post-training pruning algorithms.}
    \label{tab:comparison}
    \begin{threeparttable}
    \resizebox{\textwidth}{!}{
    \renewcommand{\arraystretch}{1.8}
\begin{tabular}{lllllll}
\toprule
{\bf Algorithm}  & \bf W? & \bf Act.? & {$\vX$} & $\vY$ & {\bf $\vS_{jk}$}\tnote{\color{blue}(a)} & \bf Comment \\
\midrule
\rowcolor{bgcolor4}
\algname{General Sym.} & \cmark & \cmark & $\vX$ & $\vY$ & $|\vW_{jk}| \left(\|\mathbf{X}_{:j}\|_2 + \|\mathbf{Y}_{k:}\|_2 \right)$ & \Cref{lemma:lm1}\\
\algname{Marginal} & \cmark & \xmark & \bf $\vI$ & \bf 0 & $|\vW_{jk}|$ & - \\
\algname{Wanda} & \cmark & \cmark & \bf X & \bf 0 & $|\vW_{jk}|\norm{\vX_{:j}}_2$ & \Cref{corollary1_2}\\
\rowcolor{bgcolor4}
\algname{OWanda}& \cmark & \cmark & \bf 0 & \bf Y & $|\vW_{jk}|\norm{\vY_{k:}}_2$ & \Cref{corollary1}\\
\rowcolor{bgcolor4}
\algname{Symmetric}& \cmark & \cmark & $\vW^T$ & $\vW^T$ & $|\vW_{jk}| \sqrt{\sqn{\vW_{j:}}_2 + \sqn{\vW_{:k}}}_2$ & \Cref{corollary2}\\

\algname{RI (v1)} & \multirow{1}{*}{\cmark} & \multirow{1}{*}{\xmark} & $t_j(1;, \cdots;, 1)$, $t_j = ({\sqrt{b} \norm{\vW_{j:}}}_1)^{-1}$\tnote{\color{blue}(a)} & $s_k(1, \cdots, 1)$, $s_k = \left({\sqrt{c}\norm{\vW_{:k}}_1}\right)^{-1}$ & \multirow{1}{*}{$\norm{\vW_{j:}}_1^{-1} + \norm{\vW_{:k}}_1^{-1}$} & \Cref{thm:main2}\\  
\algname{RI (v2)}  & \cmark & \xmark & $\Diag(\|\mathbf{W}_{1:}\|^{-1}_1, \ldots, \|\mathbf{W}_{b:}\|^{-1}_1)$ & $\Diag(\|\mathbf{W}_{:1}\|^{-1}_1, \ldots, \|\mathbf{W}_{:c}\|^{-1}_1)$ & $\norm{\vW_{j:}}_1^{-1} + \norm{\vW_{:k}}_1^{-1}$ & \Cref{thm:main2}\\

\algname{RIA} & \cmark & \cmark & $\delta_{u=j} \delta_{v=p} {\left\|\mathbf{C}_{: j}\right\|_2^\alpha}{\left\|\mathbf{W}_{j:}\right\|_1^{-1}}$\tnote{\color{blue}(c)} & $\delta_{u=s} \delta_{v=k} {\left\|\mathbf{C}_{: j}\right\|_2^\alpha}{\left\|\mathbf{W}_{:k}\right\|_1^{-1}}$ & $\left({\left\|\mathbf{W}_{j:}\right\|_1^{-1}}+{\left\|\mathbf{W}_{:k}\right\|_1^{-1}}\right) \left\|\mathbf{\vX}_{:j}\right\|_2^{\alpha}$ & \Cref{lem:ria}\\

\rowcolor{bgcolor4}
\algname{General (diag.)} & \cmark & \cmark & $\vA \vD_{\vX}$\tnote{\color{blue}(d)} & $\vD_{\vY}\vB$ & ${\left\|\mathbf{A}_{:j}\right\|_2}{\left\|\mathbf{W}_{j:}\right\|_1^{-1}} + {\left\|\mathbf{B}_{k:}\right\|_2}{\left\|\mathbf{W}_{:k}\right\|_1^{-1}}$ & \Cref{lem:general2}\\

\rowcolor{bgcolor4}
\algname{$\ell_p$-norm (v1)} & \cmark & \xmark\tnote{\color{blue}(e)} & ${\left\|\mathbf{W}_{j:}\right\|_p^{-1} \cdot\left\|\mathbf{W}_{j:}\right\|_2^{-1}} \cdot \mathbf{W}_{j:}^{\top}$ & ${\left\|\mathbf{W}_{:k}\right\|_p^{-1} \cdot\left\|\mathbf{W}_{:k}\right\|_2^{-1}} \cdot \mathbf{W}_{:k}^{\top}$ & \multirow{1}{*}{$|\mathbf{W}_{jk}| (\|\mathbf{W}_{j:}\|^{-1}_p + \|\mathbf{W}_{:k}\|^{-1}_p)$} &\Cref{lem:generalized_p_norm}\\

\rowcolor{bgcolor4}
\algname{$\ell_p$-norm (v2)} & \cmark & \xmark & $\left\|\mathbf{W}_{j:}\right\|_p^{-1} \cdot \mathbf{u}$ & $\left\|\mathbf{W}_{:k}\right\|_p^{-1} \cdot \mathbf{v}$ & $|\mathbf{W}_{jk}| (\|\mathbf{W}_{j:}\|^{-1}_p + \|\mathbf{W}_{:k}\|^{-1}_p)$ & \Cref{lem:random_unit_vector_scaling}\\

\rowcolor{bgcolor4}
\algname{StochRIA} & \cmark & \xmark & ${\mathbf{1}_{\{i \in S_j\}}}\left({\|\mathbf{W}_{j:S_j}\|_{1}\sqrt{\tau}}\right)^{-1}$ & ${\mathbf{1}_{\{i \in S_k\}}} \left({\|\mathbf{W}_{S_k:k}\|_{1}\sqrt{\tau}}\right)^{-1}$ & $|\mathbf{W}_{jk}| (\|\mathbf{W}_{j:S_j}\|_1^{-1} + \|\mathbf{W}_{S_k:k}\|_1^{-1})$ & \Cref{lem:stochria}\\
\bottomrule
\end{tabular}}
 %%%%%%%%%%%%%%%%%%%%%%%%%%
  \begin{tablenotes}
        {\tiny
        \item [{\color{blue}(a)}] Without loss of generality, we consider the elimination of a single weight, $\vW_{jk}$. The detailed explanation can be found in \Cref{lemma:lm1} and \Cref{sec:new_form}.
        % \item [{\color{blue}(b)}] All metrics having column-wise or row-wise design like $\vW_{:k}$ and $\vW_{j:}$ is the instantialization of $\vX_{:j}$ and $\vY_{k:}$ instead 
        \item [{\color{blue}(b)}] For simplicity, instead of displaying the entire matrices $\mathbf{X}$ and $\mathbf{Y}$, we present the columns $\mathbf{X}_{:j}$ and the rows $\mathbf{Y}_{k:}$. This design is employed in the algorithms \algname{RI}, \algname{RIA}, $\ell_p$-norm, and \algname{StochRIA}.
        \item [{\color{blue}(c)}] The Kronecker delta, denoted by $\delta_{i j}$, is a function of two indices $i$ and $j$ that equals 1 if $i=j$ and 0 otherwise.
        \item [{\color{blue}(d)}] $\mathbf{D}_{\mathbf{X}}$ and $\mathbf{D}_{\mathbf{Y}}$ are the diagonal matrices associated with $\mathbf{W}$, as defined in \Cref{sec:general_solution}. 
        \item [{\color{blue}(e)}] By default, for \algname{$\ell_p$-norm} and \algname{StochRIA}, we do not consider the input activation. However, the design is similar to the transition from \algname{RI} to \algname{RIA}, as described in \Cref{sec:ria}. % For simplicity and illustration purposes, we present the version without activation functions.
        }  
    \end{tablenotes}
 %%%%%%%%%%%%%%%%%%%%%%%%%%    
    \end{threeparttable}
\end{table*}

% \kai{TODO: fix the column representation in the above figure and fix the RI :k and j: difference}
  
\section{Related Work}
\paragraph{Traditional model pruning.} 
% \subsection{Traditional Model Pruning}
Pruning has emerged as a powerful strategy to compress and accelerate deep neural networks by removing redundant connections while preserving overall performance \citep{han2015learning, frankle2018lottery, hoefler2021sparsity}. Early works introduced iterative pruning-and-retraining approaches, which iteratively identify unimportant weights, discard them, and retrain the resulting sparse network to recover accuracy \citep{lecun1989optimal, han2015learning}. More recent dynamic sparse training techniques \citep{mocanu2018scalable, bellec2018deep, lee2018snip, mostafa2019parameter} start from a sparse initialization and continuously prune and grow connections throughout training. These methods integrate sparsification into the training loop, yielding promising trade-offs between model size and performance. A prominent line of work has leveraged learnable thresholds to realize non-uniform sparsity \citep{kusupati2020soft} or combined magnitude-based pruning with periodic connectivity updates to regrow valuable weights \citep{RigL, SRigL}. However, most of these methods still rely on standard back-propagation over the full parameter set, which can be prohibitively expensive when scaling up to LLMs.
 
% \subsection{LLM Post-Training Pruning}
% With the rise of LLMs, researchers have proposed methods to address the challenges of pruning these models \citep{li2023sparse}. Recent efforts focus on post-training pruning, which starts from a pre-trained network and removes redundant parameters without requiring end-to-end fine-tuning or retraining. \algname{SparseGPT} uses second-order information to solve a layer-wise reconstruction problem, enabling unstructured and N:M structured sparsity \citep{zhou2021learning}. Similarly, \algname{Wanda} \citep{Wanda} introduces a pruning metric that accounts for both weight magnitude and corresponding input activations, achieving perplexity results comparable to \algname{SparseGPT} \citep{SparseGPT}.
% However, \cite{jaiswal2023compressing} highlights that perplexity alone may not be a reliable metric for assessing model compression, noting that both \algname{SparseGPT} and \algname{Wanda} underperform at even modest sparsity levels (25-30\%). This raises concerns that the absence of retraining after pruning could result in significant performance degradation. To address this limitation, we propose a novel training method that emphasizes high training efficiency to mitigate these issues. Additionally, \algname{DSnoT} \citep{zhang2023dynamic} presents a simple yet effective approach for pruning and regrowing weights based on their values and statistical properties (e.g., expectation and variance) for each pruning row, achieving results without the need for retraining.
\paragraph{LLM post-training pruning.} 
% \subsection{LLM Post-Training Pruning}
The substantial computational demands of LLMs have raised the development of pruning methods tailored to reduce parameters counts without compromising performance \citep{li2023sparse, zhu2024survey}. Among these methods, post-training pruning eliminates redundant parameters in a pre-training network without requiring resource-intensive fine-tuning. For instance, \algname{SparseGPT} \citep{SparseGPT} leverages second-order information to solve layer-wise reconstruction problems, supporting both unstructured and N:M structured sparsity \citep{zhou2021learning}. \algname{Wanda} \citep{Wanda} introduces a pruning metric that incorporates both weight magnitudes and corresponding input activations, achieving perplexity performance comparable to \algname{SparseGPT} while surpassing simple magnitude-based pruning. The \algname{RIA} method \citep{RIA} builds on \algname{Wanda} by considering relative weight importance, offering performance improvements at minimal additional cost. Moreover, \algname{DSnoT} \citep{zhang2023dynamic} proposes pruning and regrowing weights based on statistical properties (e.g., mean and variance) in each pruning row, obviating the need for retraining. 

\section{Symmetric Wanda}
\subsection{Prerequisites}
Post-training pruning is defined as follows: consider a target sparsity ratio $\varepsilon \in [0, 1)$, a set of calibration inputs $\mathbf{X} \in \mathbb{R}^{a \times b}$, and pre-trained weights $\mathbf{W} \in \mathbb{R}^{b \times c}$. For clarity in the mathematical framework, we abstract the dimensions of inputs and weights. Specifically, in the context of large language models, let $a \eqdef C_{\text{in}}$, $b \eqdef N \times L$, and $c \equiv C_{\text{out}}$, where $N$ and $L$ denote the batch size and sequence length, respectively. The objective is to identify an optimal pruned weight matrix $\widetilde{\mathbf{W}} \in \mathbb{R}^{b \times c}$ that minimizes:

\begin{equation}\label{obj1}\tag{InpRecon}
    f(\widetilde{\mathbf{W}}) \eqdef \|\mathbf{X} (\widetilde{\mathbf{W}} - \mathbf{W})\|_F^2,
\end{equation}

where the optimization challenge is:
\begin{equation}
    \minimize f(\widetilde{\mathbf{W}}) \ \ s.t. \ \ \text{Mem}(\widetilde{\mathbf{W}}) \leq (1 - \varepsilon) \text{Mem}(\mathbf{W}), \notag
\end{equation}

% where $\text{Mem}(\cdot)$ represents the memory consumption associated with a weight matrix. (\ref{obj1}) represents the input reconstruction error. 
where $\text{Mem}(\cdot)$ denotes the memory consumption associated with a weight matrix, and (\ref{obj1}) quantifies the input reconstruction error.

This formulation applies to various post-training compression techniques, including both pruning \citep{SparseGPT, Wanda, RIA} and quantization \citep{GPTQ, AQLM}. Our focus here is specifically on post-training pruning.

\subsection{Symmetric Wanda: New Formulations}\label{sec:new_form}
Building upon the methods introduced in \algname{Wanda} \citep{Wanda}, which considered both weights and activations, and later improvements by \algname{RIA} \citep{RIA}, which analyzed the relative importance of weights by summing over corresponding rows and columns, we provide new insights by redefining our optimization objective. Apart from the previous defined input calibration $\vX$, we particularly introduce the output calibration $\vY \in \mathbb{R}^{c\times d}$. Considering both the input and output dependencies, we express the objective as:
% Building upon the methods introduced in \algname{Wanda} \citep{Wanda}, which considered both weights and activations, and later improvements by \algname{RIA} \citep{RIA}, which analyzed the relative importance of weights by summing over corresponding rows and columns, we provide new insights by redefining our optimization objective. Beyond the defined input calibration $\vX$, we particularly introduce the output calibration $\vY \in \mathbb{R}^{c\times d}$. Considering both the input and output dependencies, we express the objective as:

% \begin{equation}\label{obj2}\tag{Sym}
\vspace{-8mm}
\begin{equation}\label{obj2}\tag{Sym}
    g(\widetilde{\mathbf{W}}) \eqdef \|\mathbf{X} (\widetilde{\mathbf{W}} - \mathbf{W})\|_F + \| (\widetilde{\mathbf{W}} - \mathbf{W})\vY\|_F,
\end{equation}

and propose to solve:
\begin{equation}
    \operatorname{minimize} \ g(\widetilde{\mathbf{W}}), \ \ s.t. \ \ \text{Mem}(\widetilde{\mathbf{W}}) \leq (1 - \varepsilon) \text{Mem}(\mathbf{W}). \notag
\end{equation}

% We name the approach using the general matrix in (\ref{obj2}) without instantialization as \algname{SymWanda}, aims to minimize the reconstruction error influenced by both the input $\mathbf{X}$ and the output $\mathbf{Y}$. We elucidate the efficacy of this method and offer new theoretical insights into the performance advantages previously observed with \algname{Wanda} and \algname{RIA}.
% We refer to the approach using the general matrix in (\ref{obj2}) without instantiation as \algname{SymWanda}, which aims to minimize the reconstruction error influenced by both the input $\mathbf{X}$ and the output $\mathbf{Y}$. Noticed that here we consider the formulation without \emph{squared} Frobenius norms (we provide the version with squared Frobenius norms in \Cref{sec:squared_frobenius}). We elucidate the efficacy of this method and provide new theoretical insights into the performance advantages previously observed with \algname{Wanda} and \algname{RIA}.
We refer to the method that utilizes the general matrix in (\ref{obj2}) without instantiation as \algname{SymWanda}, which is designed to minimize the reconstruction error affected by both the input $\mathbf{X}$ and the output $\mathbf{Y}$. It is important to note that this formulation employs \emph{non-squared} Frobenius norms to facilitate better theoretical interpretations. A squared norm version is also provided in \Cref{sec:squared_frobenius} for comparison. We elucidate the efficacy of both approaches and provide new theoretical insights into the performance advantages previously observed with \algname{Wanda} and \algname{RIA}.

\begin{lemma}\label{lemma:lm1}
    Assume we aim to eliminate a single weight $\vW_{jk}$, setting $\widetilde{\mathbf{W}}_{jk} = 0$ and keeping all other weights unchanged. The simplified expression for $g(\widetilde{\mathbf{W}})$ becomes:

    \begin{equation}\label{eqn0}
        g(\widetilde{\mathbf{W}}) = |\vW_{jk}| \left(\|\mathbf{X}_{:j}\|_2 + \|\mathbf{Y}_{k:}\|_2 \right) \eqdef \vS_{jk}, 
    \end{equation}

    where $\mathbf{X}_{:j}$ and $\mathbf{Y}_{k:}$ represent the j-th column and k-th row of $\mathbf{X}$ and $\mathbf{Y}$, respectively. 
\end{lemma}

This formulation (\ref{eqn0}) underscores the impact of individual weights on the error metrics and guides the pruning process. While Lemma \ref{lemma:lm1} simplifies the formulation for pruning a single weight, the general approach can be extended to multiple weights iteratively. This method facilitates a robust pruning strategy that is backed by both empirical results and theoretical foundations, bridging the gap in understanding observed in prior studies such as \algname{Wanda} \citep{Wanda} and \algname{RIA} \citep{RIA}.

% \begin{comment}\label{thm:main1}
%     If we choose $\vX = \mathbf{0}\in \mbR^{a\times b}$, then our pruning method reduces to \emph{output} Wanda: $\vS_{jk} \eqdef |\vW_{jk}|\norm{\vY_{k:}}_2$; If we choose $\vY = \mathbf{0}\in \mbR^{a\times b}$, then our pruning method reduces to \emph{input} Wanda: $\vS_{jk} \eqdef |\vW_{jk}|\norm{\vX_{:j}}_2$. This is exactly the objective in \cite{Wanda}.
% \end{comment}

\begin{corollary}\label{corollary1_2}
    Setting $\mathbf{Y} = \mathbf{0} \in \mathbb{R}^{c \times d}$ transitions our method to \emph{input} \algname{Wanda}, described by $\mathbf{S}_{jk} \eqdef |\mathbf{W}_{jk}| \|\mathbf{X}_{:j}\|_2$. 
\end{corollary}

This directly aligns with the objective in \cite{Wanda}, demonstrating that \algname{Wanda} is a specific case under our broader framework.

\begin{corollary}\label{corollary1} 
    Conversely, choosing $\mathbf{X} = \mathbf{0} \in \mathbb{R}^{a \times b}$ simplifies our pruning method to what we term \emph{output} \algname{Wanda} (denoted as \algname{OWanda}), where the score matrix becomes $\mathbf{S}_{jk} \eqdef |\mathbf{W}_{jk}| \|\mathbf{Y}_{k:}\|_2$. 
\end{corollary}

\begin{corollary}\label{corollary2}
By setting $\mathbf{X} = \mathbf{W}^\top \in \mathbb{R}^{c \times b} (a = c)$ and $\mathbf{Y} = \mathbf{W}^\top \in \mathbb{R}^{c \times b} (d=b)$, the score matrix $\mathbf{S}_{jk}$ is redefined as $|\mathbf{W}_{jk}| (\|\mathbf{W}_{j:}\|_2 + \|\mathbf{W}_{:k}\|_2)$. 
\end{corollary} 

This configuration suggests an alternative masking approach and segues into a further analysis on how our method encompasses both \algname{Wanda} and \algname{RIA} as special cases. The following theorem provides a provable construction to recover the relative importance design in \cite{RIA}.

\begin{theorem}\label{thm:main2}
    Assuming $a = b$ and $c = d$, consider one of the following strategies:
    \begin{itemize}
        \item $\mathbf{X}_{:j} \eqdef t_{j} (1; \ldots; 1) \in \mathbb{R}^{b \times 1}$ and $\mathbf{Y}_{k:} \eqdef s_k (1, \ldots, 1) \in \mathbb{R}^{1 \times c}$, where $t_j = ({\sqrt{b} \|\mathbf{W}_{j:}\|_1})^{-1}$ and $s_k = ({\sqrt{c} \|\mathbf{W}_{:k}\|_1})^{-1}$.
        \item $\mathbf{X} = \Diag(\|\mathbf{W}_{1:}\|^{-1}_1, \ldots, \|\mathbf{W}_{b:}\|^{-1}_1)$ and $\mathbf{Y} = \Diag(\|\mathbf{W}_{:1}\|^{-1}_1, \ldots, \|\mathbf{W}_{:c}\|^{-1}_1)$.
    \end{itemize}
    For these configurations, the condition $\|\mathbf{X}_{:j}\|_2 + \|\mathbf{Y}_{k:}\|_2 = \alpha_{jk} \eqdef {\|\mathbf{W}_{j:}\|_1^{-1}} + {\|\mathbf{W}_{:k}\|_1^{-1}}$ holds for all $j, k$.
\end{theorem}

This theorem elucidates that our methodology can invariably reconstruct the framework of relative importance \algname{RI} in \citep{RIA}, validating the adaptability and breadth of our proposed pruning strategy. 
% Noticed that here for the first design in \Cref{thm:main2}, for simplexity we provide the explicit formulation of the $j$-th row of $\vX$

\subsection{From Relative Importance (RI) to RI Activation}\label{sec:ria}
In \Cref{thm:main2}, we revisit the concept of Relative Importance (\algname{RI}). Specifically, we represent \algname{RI} by the following equation:
$$
\vS_{jk} = {|\vW_{jk}|}{\norm{\vW_{j:}}_1^{-1}} + {|\vW_{jk}|}{\norm{\vW_{:k}}_1^{-1}} \eqdef \algname{RI}_{jk}.
$$

\cite{RIA} also introduces an enhanced version of \algname{RI}, termed RI with Activation (\algname{RIA}), which incorporates the $\ell_2$-norm of activations:

\begin{equation}\label{eqn:ria}
    \algname{RIA}_{jk} = \algname{RI}_{jk} \cdot \norm{\vX_{:j}}_2^\alpha,
\end{equation}

where \(\alpha\) is controlling the strength of activations.

This section aims to explore the derivation of \(\algname{RIA}\) with theoretical grounding in \(\algname{RI}\). To clarify our notation and avoid confusion, we are aiming at finding the suitable $\vA \in \mathbb{R}^{a\times b}$ and $\vB \in \mathbb{R}^{c\times d}$ such as:
$$
\left\|\mathbf{A}_{j:}\right\|_2+\left\|\mathbf{B}_{:k}\right\|_2=\left({\left\|\mathbf{W}_{j:}\right\|_1^{-1}}+{\left\|\mathbf{W}_{:k}\right\|_1^{-1}}\right) \cdot\left\|\mathbf{C}_{:j}\right\|_2^{\alpha},
$$
where $\vC_{:j}$ will be instantiated as $\vX_{:j}$ to satisfy \Cref{eqn:ria}.

% \begin{lemma}\label{lem:ria}
%     Define \(p\) as any valid column index for \(\vA\), and set:
%     $$
%     \vA_{uv}= \begin{cases}\frac{1}{\left\|\mathbf{W}_{j:}\right\|_1} \times\left\|\mathbf{C}_{:j}\right\|_2^\alpha & \text { if } u=j, v=p \\ 0 & \text { otherwise. }\end{cases}
%     $$
    
%     Similarly, define \(s\) as any valid row index for \(\vB\), and let:
%     $$
%     \vB_{uv}= \begin{cases}\frac{1}{\left\|\mathbf{W}_{:k}\right\|_1} \times\left\|\mathbf{C}_{:j}\right\|_2^\alpha & \text { if } u=s, v=k \\ 0 & \text { otherwise. }\end{cases}
%     $$
%     We recover the relative importance activation in \Cref{eqn:ria}.
% \end{lemma}
\begin{lemma}\label{lem:ria}
Let $p$ be a valid column index for $\mathbf{A}$. Define $\mathbf{A}_{uv} = 0$ for all $(u,v)\neq (j,p)$, and 
$
\mathbf{A}_{j,p} 
= {\|\mathbf{C}_{:j}\|_2^\alpha}{\|\mathbf{W}_{j:}\|_1^{-1}}.
$
Similarly, let $s$ be a valid row index for $\mathbf{B}$. Define $\mathbf{B}_{uv} = 0$ for all $(u,v)\neq (s,k)$, and 
$
\mathbf{B}_{s,k}
= {\|\mathbf{C}_{:j}\|_2^\alpha}{\|\mathbf{W}_{:k}\|_1^{-1}}.
$
Then we recover \Cref{eqn:ria}.
\end{lemma}

The nonzero element in \(\mathbf{A}\) ensures that the \(\ell_2\)-norm of the \(j\)-th row of \(\mathbf{A}\) is:
$
\left\|\mathbf{A}_{j:}\right\|_2={\left\|\mathbf{W}_{j:}\right\|_1^{-1}} \cdot \left\|\mathbf{C}_{:j}\right\|_2^\alpha.
$
Similarly, the nonzero element in \(\mathbf{B}\) ensures that the \(\ell_2\)-norm of the \(k\)-th column of \(\mathbf{B}\) is:
$
\left\|\mathbf{B}_{:k}\right\|_2={\left\|\mathbf{W}_{:k}\right\|_1^{-1}} \cdot\left\|\mathbf{C}_{:j}\right\|_2^\alpha.
$
Combining these norms fulfills the intended equation.

\subsection{General Solution}\label{sec:general_solution}
In \Cref{thm:main2}, we presented two distinct strategies for recovering the relative importance as described in \cite{RIA}. Following this, in \Cref{lem:ria}, we constructed a method that accounts for both the weights and the input activations. Inspired by the diagonal design in \Cref{thm:main2}, we now propose a general variant that considers both the weights and the activations.

Given that $\mathbf{D}_{\mathbf{X}} \in \mathbb{R}^{b\times b}$ and $\mathbf{D}_{\mathbf{Y}} \in \mathbb{R}^{c\times c}$ are diagonal matrices with entries defined as $\left(\mathbf{D}_{\mathbf{X}}\right)_{ii} = x_i = \left\|\mathbf{W}_{i:}\right\|_1^{-1}$ and $\left(\mathbf{D}_{\mathbf{Y}}\right)_{ii} = y_i = \left\|\mathbf{W}_{:i}\right\|_1^{-1}$ respectively, and $\mathbf{A}\in \mathbb{R}^{a\times b}$ and $\mathbf{B}\in \mbR^{c\times d}$ are arbitrary matrices, our objective is to compute the sum of norms:
$
\left\|\left(\mathbf{A} \mathbf{D}_{\mathbf{X}}\right)_{:j}\right\|_2 + \left\|\left(\mathbf{D}_{\mathbf{Y}} \mathbf{B}\right)_{k:}\right\|_2.
$

\begin{lemma}\label{lem:general2}
    Given the above definition, we show 
    $$
    \left\|\left(\mathbf{A} \mathbf{D}_{\mathbf{X}}\right)_{:j}\right\|_2 + \left\|\left(\mathbf{D}_{\mathbf{Y}} \mathbf{B}\right)_{k:}\right\|_2 = \frac{\left\|\mathbf{A}_{:j}\right\|_2}{\left\|\mathbf{W}_{j:}\right\|_1} + \frac{\left\|\mathbf{B}_{k:}\right\|_2}{\left\|\mathbf{W}_{:k}\right\|_1}.
    $$
\end{lemma}

The utilization of the diagonal matrices $\mathbf{D}_{\mathbf{X}}$ and $\mathbf{D}_{\mathbf{Y}}$ simplifies the sum of the norms to the expressions derived above, offering insights into the influence of the weight matrix $\mathbf{W}$ on the norms of matrix transformations.

\subsection{Enhanced Relative Importance Strategies}\label{sec:strategies}
Beyond \algname{RIA}, we propose several alternative strategies for relative importance that aim to minimize $\vS_{jk}$ in \Cref{eqn0}.

\subsubsection{Generalized $\ell_p$-Norm}\label{sec:p_norm}
Expanding beyond the conventional $\ell_1$-norm, we explore the utility of the $\ell_p$-norm in designing score matrices. %We propose two variations:
% $\bullet$ In one approach, the score matrix is weighted directly by the weights, analogous to the method discussed in \Cref{corollary1}. The score for each weight is calculated as:
% \begin{equation}\label{eqn:pnorm1}
%     \mathbf{S}_{jk} = |\mathbf{W}_{jk}| (\|\mathbf{W}_{j:}\|_p + \|\mathbf{W}_{:k}\|_p).
% \end{equation}
% $\bullet$ In another, mirroring the strategy outlined in Theorem \ref{thm:main2} for reconstructing \algname{RIA} outcomes, we define the score as:
In our approach, mirroring the strategy outlined in Theorem \ref{thm:main2} for reconstructing \algname{RIA} outcomes, we define the score as:

\begin{equation}\label{eqn:pnorm2}
    \mathbf{S}_{jk} = |\mathbf{W}_{jk}| (\|\mathbf{W}_{j:}\|^{-1}_p + \|\mathbf{W}_{:k}\|^{-1}_p).
\end{equation}

Next, we are interested in finding the explicit formulation of $\vX$ and $\vY$ instead of the norm representation when constructing the general $\ell_p$-norm. 

% \begin{theorem}[Generalized p-norm]\label{thm:generalized_p_norm}
%     If $\mathbf{X}_{: j}=\frac{\left\|\mathbf{W}_{j:}\right\|_p}{\left\|\mathbf{W}_{j:}\right\|_2} \cdot \mathbf{W}_{j:}^{\top} \cdot$ and $\mathbf{Y}_{k:}=\frac{\left\|\mathbf{W}_{: k}\right\|_p}{\left\|\mathbf{W}_{: k}\right\|_2} \cdot \mathbf{W}_{: k}^{\top}$; or 
% \end{theorem}

\begin{lemma}[Generalized $\ell_p$-norm]\label{lem:generalized_p_norm}
    % Let $\mathbf{X}_{: j}=\frac{\left\|\mathbf{W}_{j:}\right\|_p}{\left\|\mathbf{W}_{j:}\right\|_2} \cdot \mathbf{W}_{j:}^{\top}$ and $\mathbf{Y}_{k:}=\frac{\left\|\mathbf{W}_{: k}\right\|_p}{\left\|\mathbf{W}_{: k}\right\|_2} \cdot \mathbf{W}_{: k}^{\top}$, we show \Cref{eqn:pnorm1}.
    
    Let $\mathbf{X}_{: j}={\left\|\mathbf{W}_{j:}\right\|_p^{-1} \cdot\left\|\mathbf{W}_{j:}\right\|_2^{-1}} \cdot \mathbf{W}_{j:}^{\top}$ and $\mathbf{Y}_{k:}={\left\|\mathbf{W}_{:k}\right\|_p^{-1} \cdot\left\|\mathbf{W}_{:k}\right\|_2^{-1}} \cdot \mathbf{W}_{:k}^{\top}$, we recover \Cref{eqn:pnorm2}.
\end{lemma}

% We have alternative choices to construct \Cref{eqn:pnorm2}. For simplexity, we focus on \Cref{eqn:pnorm2} is almost the same by considering the negative exponent, which is almost staightforward. 

Since the equation only requires $\left\|\mathbf{X}_{: j}\right\|_2=\left\|\mathbf{W}_j\right\|_p^{-1}$, \emph{any} vector with this $\ell_2$-norm will satisfy the condition. Inspired by this fact, we can consider the random unit vector scaling in the below lemma. 

\begin{lemma}[Random unit vector scaling]\label{lem:random_unit_vector_scaling}
    Choose any unit vector $\mathbf{u}, \mathbf{v}$ (i.e., $\|\mathbf{u}\|_2=1, \norm{\mathbf{v}}_2=1$) and set $\mathbf{X}_{: j}=\left\|\mathbf{W}_{j:}\right\|_p^{-1} \cdot \mathbf{u}$ and $\mathbf{Y}_{k:}=\left\|\mathbf{W}_{:k}\right\|_p^{-1} \cdot \mathbf{v}$ ensuring \Cref{eqn:pnorm2}.
\end{lemma}

% This lemma shows that for any unit vector, it is possible to construct feasible matrices that satisfy \Cref{eqn:pnorm2}.

\subsubsection{Stochastic Relative Importance}\label{sec:StochRIA}
Considering the computational and noise challenges associated with summing all elements across the full rows and columns of large matrices, we introduce a stochastic approach that involves sampling a subset of each row and column. This method assesses the effects of varying subset sizes, denoted by $\tau$, where $\tau < \min(b, c)$, on the overall performance. Specifically, we aim to:

a) Evaluate the sensitivity of the final performance to the size of $\tau$ when $\tau$ is reasonably large.

b) Determine if random sampling can enhance the results compared to a deterministic approach.

For this, we define the score matrix for a randomly sampled subset as:
\begin{equation}\label{eqn:stochria}
    \mathbf{S}_{jk} = |\mathbf{W}_{jk}| (\|\mathbf{W}_{j:S_j}\|_1^{-1} + \|\mathbf{W}_{S_k:k}\|_1^{-1}),
\end{equation}

where $S_j$ and $S_k$ represent the sampled indices from the $j$-th row and $k$-th column, respectively, each with a cardinality of $\tau$. This approach builds on the \algname{RIA}-inspired framework, adapting it for practical scenarios involving large-scale data.

For \algname{RIA} in each weight layer, the reweighting sampling complexity is $O(b + c)$. In LLMs, $b$ and $c$ are always very large. Let's say the selection ratio is $\beta$, then for the stochastic relative importance design, the sampling complexity can be reduced to $O(\beta \min (b, c))$, which has been highly reduced. 

% \begin{lemma}\label{lem:stochria}
%     Let 
%     $
%     \begin{cases}
%         \dfrac{1}{\left\| \mathbf{W}_{j:S_j} \right\|_1} \cdot \dfrac{1}{\sqrt{\tau}} & \text{if } i \in S_j, \\
%         0 & \text{otherwise}.
%     \end{cases}
%     $  
%     $
%     \begin{cases}
%         \dfrac{1}{\left\| \mathbf{W}_{S_k:k} \right\|_1} \cdot \dfrac{1}{\sqrt{\tau}} & \text{if } i \in S_k, \\
%         0 & \text{otherwise}.
%     \end{cases}
%     $
%     recovers \Cref{eqn:stochria}.
% \end{lemma}

% \begin{lemma}\label{lem:stochria}
% Let $S_j$ and $S_k$ be given index sets, and let $\tau > 0$. Define the vectors $\vX_{:j}$ and $\vY_{k:}$ entrywise by
% $$
% \vX_{:j}(i) =
% \begin{cases}
% \dfrac{1}{\|\mathbf{W}_{j:S_j}\|_{1}\sqrt{\tau}}, & \text{if } i \in S_j,\\[8pt]
% 0, & \text{otherwise},
% \end{cases}
% $$

% $$
% \vY_{k:}(i) =
% \begin{cases}
% \dfrac{1}{\|\mathbf{W}_{S_k:k}\|_{1}\sqrt{\tau}}, & \text{if } i \in S_k,\\[8pt]
% 0, & \text{otherwise}.
% \end{cases}
% $$
% Then these vectors satisfy \Cref{eqn:stochria}.
% \end{lemma}
\begin{lemma}\label{lem:stochria}
Let $S_j$ and $S_k$ be index sets, and let $\tau > 0$. Define the vectors $\vX_{:j}$ and $\vY_{k:}$ by
$$
\vX_{:j}(i) 
= \frac{\mathbf{1}_{\{i \in S_j\}}}{\|\mathbf{W}_{j:S_j}\|_{1}\sqrt{\tau}},
\quad
\vY_{k:}(i) 
= \frac{\mathbf{1}_{\{i \in S_k\}}}{\|\mathbf{W}_{S_k:k}\|_{1}\sqrt{\tau}}.
$$
Then these vectors satisfy \Cref{eqn:stochria}.
\end{lemma}

\subsection{Training-Free Fine-Tuning}\label{sec:training_free_fine_tuning}
We explore training-free fine-tuning within the context of the pruning-and-growing framework. Specifically, for the pruned weight matrix $\widetilde{\vW}$, we aim to minimize the reconstruction error as defined in (\ref{obj2}). Initially, we identify the growth index, followed by the pruning index, to maintain a consistent sparsity ratio. \algname{DSnoT} \citep{zhang2023dynamic} developed a growing criterion based on the expected change in reconstruction error when reinstating a weight. Particularly, for any given weight row $q\in [1, b]$, the index $i$ is determined as follows:
$$
i = \argmax_r \ \mathrm{sign}(\ec{\epsilon_q}) \cdot \widetilde{\vW}_{q, r} \cdot {\ec{\vX_q}}/{\mathrm{Var}(\vX_q)},
$$

where $\epsilon_q \coloneqq \vW_{q:}\vX - \widetilde{\vW}_{q:}\vX$ denotes the reconstruction error of the $q$-th row across different input activations. It is important to note that for simplicity, output activations are not considered here, which may provide an interesting avenue for future exploration. The functions $\mathrm{sign}(\cdot)$, $\ec{\cdot}$, and $\mathrm{Var}(\cdot)$ denote the standard sign function, expectation, and variance of given inputs over $N \times L$ tokens, respectively. Drawing inspiration from the \algname{Wanda} metric, the \algname{DSnoT} model defines the pruning index $j$ as:
$$
j=\underset{r: \Delta(q, r)<0}{\arg \min } |\widetilde{\vW}_{q, r}|\left\|\vX_q\right\|_2,
$$
where $\Delta(q, r) \eqdef \mathrm{sign}\!\bigl(\ec{\epsilon_q}\bigr)\,\bigl(\widetilde{\vW}_{q, r}\cdot\ec{\vX_q}\bigr)$. 

Several simple yet effective modifications have been incorporated into the pruning-and-growing framework:

\textbf{a) Relative weight importance.} Both in determining the growing index $i$ and the pruning index $j$, we incorporate global information, emphasizing the relative importance of weights in neuron selection.

\textbf{b) Square root activation.} Our follow-up experiments on \algname{Wanda} and \algname{RIA} demonstrate the benefits of square root activation in determining the pruning index $j$.
 
\textbf{c) Regularized objective.} The method \algname{MagR} \citep{zhang2024magr} found that adding an $\ell_{\infty}$ norm helps reduce the magnitude of weights during quantization. Here, we adopt a more general regularizer, considering a general $\ell_p$ norm and focusing on specific rows rather than entire layers to reduce communication costs.

% Define $\vD_{q, r}\eqdef \left(\left\|\widetilde{\mathbf{W}}_{q,:}\right\|_1^{-1}+\left\|\widetilde{\mathbf{W}}_{:, r}\right\|_1^{-1}\right)^{-1}$. The updated rule for identifying the growing index $i$ is formalized as:
Define $\vD_{q, r}\eqdef \|\widetilde{\mathbf{W}}_{q,:}\|_1^{-1}+\|\widetilde{\mathbf{W}}_{:, r}\|_1^{-1}$. The updated rule for identifying the growing index $i$ is formalized as:

\begin{equation}\label{eqn:ft01}
\begin{aligned}
    i &= \argmax_r \left\{ \mathrm{sign}(\mathbb{E}[\epsilon_q]) \cdot \vD_{q, r} \cdot \frac{\mathbb{E}[\mathbf{X}_q]}{\mathrm{Var}(\mathbf{X}_q)} + \gamma_1 \|\widetilde{\mathbf{W}}_q\|_p \right\},
\end{aligned}
\end{equation}

where $\gamma_1$ is the growing regularization parameter, striking a balance between fidelity and the $\ell_p$ regularizer. Similarly, the pruning index $j$ is now defined as:

\begin{equation}\label{eqn:ft02}
\begin{aligned}
    j&=\underset{r: \Delta(q, r)<0}{\arg \min }\left\{ |\widetilde{\vW}_{q, r}| \cdot \vD_{q,r} \cdot \left\|\vX_q\right\|_2^{\alpha} + \gamma_2 \|\widetilde{\vW}_q\|_p \right\}, 
\end{aligned}
\end{equation}

where $\Delta(q, r) \eqdef \mathrm{sign}!\bigl(\ec{\epsilon_q}\bigr),\left(\widetilde{\vW}{q, r} \cdot \vD{q, r} \cdot \ec{\vX_q}\right)$, and $\gamma_2$ denotes the pruning regularization parameter.

We name this approach \emph{Relative and Regularized Dynamic Sparse No Training} (\ft). It enables efficient network fine-tuning without additional training, conserving computational resources while enhancing performance.

\section{Experiments}
\paragraph{Setup and configurations.}
We assess the proposed methods across a broad spectrum of popular LLMs, including LlaMA2 (7b-13b) \citep{LlaMA2}, LlaMA3-8b \citep{LlaMA3}, OPT-1.3b \citep{OPT}. We utilize publicly available model checkpoints from the HuggingFace Transformers library \citep{wolf2020transformers} for our evaluations. Each experiment, focused on post-training pruning, is conducted on an NVIDIA A100-80G GPU.
The effectiveness of each pruned model is primarily measured using the perplexity score on the Wikitext-2 dataset \citep{WikiText2}. 
For calibration, we use 128 samples from the C4 dataset \citep{C4}, with each sample comprising 2048 tokens. This approach ensures consistency with the settings used in baseline methods, enabling a fair comparison.

\subsection{Efficiency of Stochastic Methods}\label{sec:efficiency_stochastic_methods}
We begin by examining two key designs discussed in \Cref{sec:strategies}: the generalized $\ell_p$ norm and stochastic relative importance. The results for the $\ell_p$ norm are presented in \Cref{sec:optimal_p}, where we confirm that $p = 1$ is indeed optimal. We also compare various $\ell_p$ norm reweighting strategies, with the results presented in \Cref{sec:norm_p_reweighting}. Our primary focus, however, is on the findings related to stochastic relative importance, which, to the best of our knowledge, represents the first approach to incorporating stochasticity into LLM post-training pruning.  

We analyze the impact of stochastic relative importance, with the results summarized in \Cref{tab:stoch_res}. The \algname{stochRIA} results correspond to a sampling ratio of $\beta = 0.1$. Each reported value represents the mean performance across five trials with different random seeds. Notably, even with less than only 10\% of the samples used to estimate relative importance, the results remain sufficiently representative, leading to promising outcomes.  

In addition to unstructured pruning with a sparsity ratio of $0.5$, we also explore structured pruning using the N:M pattern \citep{zhou2021learning, zhang2022learning}. The results are presented in \Cref{tab:stoch_res}. Noticed that here for intuitive comparison between \algname{RIA} and \algname{stochRIA}, we use the plain N:M structural pruning without channel permutation. These results consistently demonstrate the benefits and efficiency of our proposed method, \algname{stochRIA}. 

\begin{table}[!tb]
    \centering
    \caption{Comparison of \algname{StochRIA} ($\beta=0.1$) and \algname{RIA} on the Wikitext-2 dataset, using perplexity scores with $\alpha=1$. For \algname{StochRIA}, the mean perplexity over 5 trials is shown in dark, with standard deviation in {\green green}. Improvements and declines relative to \algname{RIA} are indicated in {\color{blue} blue} and {\red red}, respectively.}
    \label{tab:stoch_res}
    \resizebox{0.48\textwidth}{!}{
    \renewcommand{\arraystretch}{1.2}
    \begin{tabular}{l|lcccccc}
    % Method & LlaMA2-7b & LlaMA2-13b & LlaMA3-8b & OPT-1.3b\\ \hline
    % Dense  & 5.47 & 4.88 & 6.14 & 14.62 \\ \hline
    % SparseGPT & \\
    % Wanda & 7.79 & 6.28 & 10.81 & 22.19\\
    % \algname{RIA} & 6.88 & 5.95 & 9.44 & 18.94\\
    % \algname{RIA} $(\tau=0.1)$ & 6.91 & 5.95 & 9.46 & 18.78\\ \hline
    \toprule
    Sparsity & Method & Sampling & LlaMA2-7b & LlaMA2-13b & LlaMA3-8b & OPT-1.3b\\ \midrule
    - & Dense & - & 5.47 & 4.88 & 6.14 & 14.62 \\ \midrule
    \multirow{4}{*}{50\%} & \algname{Magnitude} & - & 16.03 & 6.83 & 205.44 & 1712.39\\
    % ~ & \algname{SparseGPT} & 7.02 & 6.02 & 9.36 & 17.46\\
    ~ & \algname{Wanda} & - & 7.79 & 6.28 & 10.81 & 22.19\\ \cmidrule{2-7}
    ~ & \algname{RIA} & Full & 6.88 & 5.95 & 9.44 & 18.94\\
    % \rowcolor{bgcolor4}
    ~ & \cellcolor{bgcolor4}\algname{stochRIA} & \cellcolor{bgcolor4}$10\%$ & \cellcolor{bgcolor4} $6.91_{{\color{red} -0.03}}^{{\green \pm 0.0032}}$ &  \cellcolor{bgcolor4} $5.95_{{\color{blue} +0}}^{{\green \pm 0.0033}}$ &  \cellcolor{bgcolor4} $9.46_{{\color{red} -0.02}}^{{\green \pm 0.025}}$ &  \cellcolor{bgcolor4} $18.78_{{\color{blue} +0.16}}^{{\green \pm 0.050}}$\\ \midrule
    \multirow{2}{*}{2:4} 
    % & \algname{Magnitude} & - & 37.77 & 8.89 & 2401.31 & 427.14\\
    % ~ & \algname{SparseGPT} & 10.82 & 8.84 & 16.17 & 23.89\\
    % ~ & \algname{Wanda} & - & 10.86 & 8.16 & 23.06 & 26.52\\ \cmidrule{2-7}
      & \algname{RIA} & Full & 11.31 & 8.40 & 22.89 & 27.43\\
    % \rowcolor{bgcolor4}
    ~ & \cellcolor{bgcolor4}\algname{stochRIA} & \cellcolor{bgcolor4}$10\%$ & \cellcolor{bgcolor4} $11.41_{{\color{red} -0.10}}^{{\green \pm 0.046}}$ & \cellcolor{bgcolor4} $8.44_{{\color{red} -0.04}}^{{\green \pm 0.016}}$ & \cellcolor{bgcolor4} $23.74_{{\color{blue} +0.15}}^{{\green \pm 0.230}}$ & \cellcolor{bgcolor4} $26.78_{{\color{blue} +0.65}}^{{\green \pm 0.127}}$\\ \midrule
    \multirow{2}{*}{4:8} 
    % & \algname{Magnitude} & - & 15.91 & 7.32 & 181.50 & 240.11\\
    % ~ & \algname{SparseGPT} & 8.49 & 7.02 & 12.13 & 20.34\\
    % ~ & \algname{Wanda} & - & 8.16 & 6.68 & 13.65 & 21.37\\ \cmidrule{2-7}
    ~ & \algname{RIA} & Full & 8.39 & 6.74 & 13.77 & 21.59\\
    % \rowcolor{bgcolor4}
    ~ & \cellcolor{bgcolor4}\algname{stochRIA} & \cellcolor{bgcolor4}$10\%$ & \cellcolor{bgcolor4} $8.44_{{\color{red} -0.05}}^{{\green \pm 0.014}}$ & \cellcolor{bgcolor4} $6.74_{{\color{blue} +0}}^{{\green \pm 0.013}}$ & \cellcolor{bgcolor4} $13.93_{{\color{red} -0.16}}^{{\green \pm 0.095}}$ & \cellcolor{bgcolor4} $21.49_{{\color{blue} +0.10}}^{{\green \pm 0.089}}$\\ \bottomrule
    \end{tabular}}
\end{table}

\begin{table*}[!tb]
\centering
\caption{Perplexity scores on Wikitext-2, accounting for various norm $\alpha$ values and column \& row sensitivity, with a sparsity ratio $50\%$.}\label{tab:norm_alpha_col_row}
\resizebox{0.98\textwidth}{!}{
\begin{tabular}{l|cccc|cccc|cccc|cccc}
\toprule
% \rowcolor{gray!30}
{Model} & \multicolumn{4}{c|}{{LlaMA2-7b}} & \multicolumn{4}{c|}{{LlaMA2-13b}} & \multicolumn{4}{c|}{{LlaMA3-8b}} & \multicolumn{4}{c}{{OPT-1.3b}} \\ \midrule
% \rowcolor{gray!30}
\textbf{$\alpha$}        & {0} & {0.5} & {1} & {2} & {0} & {0.5} & {1} & {2} & {0} & {0.5} & {1} & {2} & {0} & {0.5} & {1} & {2} \\
\midrule
Dense       & \multicolumn{4}{c}{ 5.47} & \multicolumn{4}{c}{ 4.88} & \multicolumn{4}{c}{ 6.14} & \multicolumn{4}{c}{ 14.62}\\ \midrule
\algname{Wanda}       & 16.03 & 7.60 & 7.79 & 8.66 & 6.83 & 6.17 & 6.28 & 7.15 & 205.44 & 10.66 & 10.81 & 12.98 & 1712.39 & 22.14 & 22.19 & 24.74 \\
\algname{Col-Sum}    & 11.59 & \color{blue} 6.83 & 6.91 & 7.46 & 6.39 & \bf \color{blue} 5.87 & 5.96 & 6.55 & 59.41 & 9.53 & 9.69 & 12.01 & 1062.66 & \color{blue} 18.28 & 18.41 & 22.25 \\
\algname{Row-Sum}    & 14.93 & 7.49 & 7.51 & 8.01 & 6.74 & 6.13 & 6.24 & 7.01 & 17.80 & 10.50 & 10.55 & 11.79 & 141.92 & 22.09 & 22.47 & 26.62 \\
\algname{RIA}         & 7.39 & \bf \color{blue} 6.81 & 6.88 & 7.37 & 5.95 & \color{blue} 5.93 & 5.95 & 6.56 & 12.07 & \bf \color{blue} 9.34 & \color{blue} 9.44 & 10.67 & 64.70 & \bf \color{blue} 18.08 & 18.94 & 23.39 \\
\bottomrule
\end{tabular}}
\end{table*}

% Moreover, when aggregating results across all examined models and baseline methods, \algname{stochRIA} achieves an accumulated perplexity that is 0.66 lower than \algname{RIA}, demonstrating that stochastic design can yield superior performance. %We attribute this improvement to the vast number of parameters in LLMs, which may benefit from sampling a representative subset rather than reweighting the entire row or column, thereby avoiding the dilution of salient features.  
% In our experiments, randomly sampling 10\% of the rows and columns consistently outperforms the importance-based selection method (summing over rows/columns), likely because it preserves the diversity needed to handle subpopulations that rely on lower-average-importance weights while also providing an implicit regularization effect that mitigates overfitting.
Furthermore, when aggregating results across all examined models and baselines, \algname{stochRIA} achieves an accumulated perplexity that is 0.66 lower than \algname{RIA}, demonstrating the effectiveness of a stochastic design. %We attribute this improvement to the large parameter space of LLMs, where sampling a representative subset rather than reweighting entire rows or columns helps avoid diluting salient features. 
This stochastic sampling preserves the diversity needed to handle subpopulations that rely on lower-average-importance weights while also helping preserve generalization by avoiding the dilution of salient features.

We also evaluate the performance across different sampling ratios, as shown in \Cref{sec:sampling_ratios}. Our main takeaway is that \algname{stochRIA} exhibits stable and competitive performance relative to \algname{RIA}, particularly when the sampling ratio $\tau \geq 0.05$. At or above this threshold, the performance remains robust and occasionally surpasses less noisy sampling configurations. However, at an extremely low sampling ratio of $\tau = 0.01$, a significant performance drop is observed. Consequently, we adopt $\tau = 0.1$ as the default setting for our experiments.

% \begin{table}[!tb]
%     \centering
%     \resizebox{0.5\textwidth}{!}{
%     \begin{tabular}{l|ccccccc}
%     Method & Sparsity & LlaMA2-7b & LlaMA2-13b & LlaMA3-8b & OPT-1.3b\\ \hline
%     Dense  & - & 5.47 & 4.88 & 6.14 & 14.62 \\ \hline
%     Wanda & 2:4 &\\
%     \algname{RIA} & 2:4 &\\
%     \algname{RIA} & 2:4\\ \hline
%     Wanda & 4:8 &\\
%     \algname{RIA} & 4:8 &\\
%     \algname{RIA} & 4:8&\\ \hline
%     \end{tabular}}
%     \caption{Perplexity scores on Wikitext-2 for structural N:M pattern.}
%     \label{tab:nm_structural_pruning}
% \end{table}

% \subsection{Additional Key Observations}
\subsection{Insights on Sensitivity, Activation, and Sparsity}
\paragraph{Column and row sensitivity.}
Compared with the \algname{Wanda} design, \algname{RIA} accounts for the relative importance of both rows and columns. However, it remains unclear whether columns and rows contribute equally to \algname{RIA}'s performance improvements. To investigate this, we conducted an extensive analysis of the significance of column-wise and row-wise relative importance, with the results shown in \Cref{tab:norm_alpha_col_row}. A key finding is that the sum of the columns has more impact on performance, indicating greater importance.

\begin{figure}[!tb]
    \centering
    \includegraphics[width=0.95\linewidth, trim = 88 40 100 68, clip]{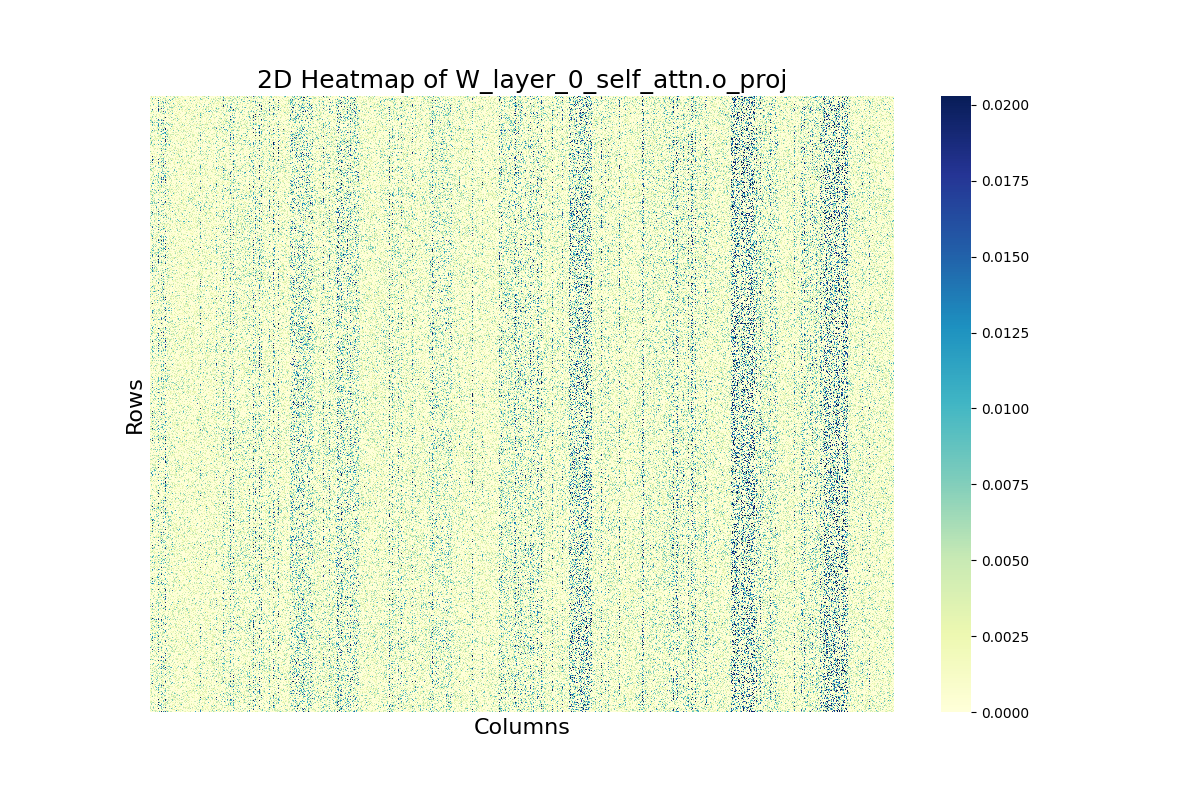}
    \vspace{-3mm}
    \caption{Visualization of the dense weight matrix in LLaMA2-7b.}
    \label{fig:vis}
\end{figure}

To provide further insights, we visualized the heatmap of a randomly selected dense weight matrix from LLaMA2-7b, as illustrated in \Cref{fig:vis}. The heatmap displays stripe-like patterns, indicating column-specific structures where certain columns show significantly higher activations, forming distinct stripes. This observation suggests that normalizing by rows effectively balances these disparities. In cases where the rows within a specific column already exhibit relatively uniform distributions, normalization over rows may not be necessary. Thus, column normalization alone might suffice to balance the contributions of output neurons, especially when some columns dominate due to large absolute values.

% \kai{TODO: add the figure showing the pattern of weights row and column.}

\paragraph{Benefits of square root input activation.} 
In the design of \algname{Wanda} \citep{Wanda}, the power factor $\alpha$ applied to input activations is set to 1, whereas in \algname{RIA} \citep{RIA}, $\alpha$ is adjusted to $0.5$.  In this study, we systematically explore the impact of varying the power factor on input activations, with detailed results presented in \Cref{tab:norm_alpha_col_row}. An $\alpha$ value of 0 implies that no activation is considered in generating the pruning matrix. Our findings consistently show that incorporating input activation improves performance in terms of perplexity. Notably, $\alpha = 0.5$ proved optimal across various methods, underscoring the advantages of reducing the magnitude of input activations. We attribute this improvement to the mitigation of outliers in the input activations, where smoothing these values provides more meaningful guidance for pruning.

\begin{table}[!tb]
    \centering
    \caption{Perplexity on Wikitext-2 with different sparsity. $\alpha=1.0$.}
    \label{tab:various_sparsity}
    \resizebox{0.48\textwidth}{!}{
    \begin{tabular}{l|ccccccc}
    \toprule
    Sparsity & Method & Sampling & L2-7b & L2-13b & L3-8b & OPT-1.3b\\ \midrule
    Dense  & - & - & 5.47 & 4.88 & 6.14 & 14.62 \\ \hline
    \multirow{3}{*}{$50\%$} & \algname{Wanda} & - & 7.79 & 6.28 & 10.81 & 22.19\\
    ~ & \algname{RIA} & Full & \color{blue} \bf 6.88 & \color{blue} \bf 5.95 & \color{blue} \bf 9.44 & 18.94\\
    % \rowcolor{bgcolor4}
    ~ & \cellcolor{bgcolor4}\algname{stochRIA} & \cellcolor{bgcolor4}$10\%$ & \cellcolor{bgcolor4}6.91 & \cellcolor{bgcolor4}\color{blue} \bf 5.95 & \cellcolor{bgcolor4}9.46 & \cellcolor{bgcolor4}\color{blue} \bf 18.78\\ \midrule
    \multirow{3}{*}{$60\%$} & \algname{Wanda} & -  & 15.30 & 9.63 & 27.55 & 38.81\\
    ~ & \algname{RIA} & Full & \color{blue} \bf 10.39 & \color{blue} \bf 7.84 & 19.52 & 26.22\\
    % \rowcolor{bgcolor4}
    ~ & \cellcolor{bgcolor4}\algname{stochRIA} & \cellcolor{bgcolor4}$10\%$ & \cellcolor{bgcolor4}10.62 & \cellcolor{bgcolor4}7.97 & \cellcolor{bgcolor4}\color{blue} \bf 19.04 & \cellcolor{bgcolor4}\color{blue} \bf 25.93\\\midrule
    \multirow{3}{*}{$70\%$} & \algname{Wanda} & -  & 214.93 & 104.97 & 412.90 & 231.15\\ 
    ~ & \algname{RIA} & Full & \color{blue} \bf 68.75 & \color{blue} \bf 51.96 & 169.51 & 98.52\\ 
    % \rowcolor{bgcolor4}
    ~ & \cellcolor{bgcolor4}\algname{stochRIA} & \cellcolor{bgcolor4}$10\%$ & \cellcolor{bgcolor4}72.85 & \cellcolor{bgcolor4}62.15 & \cellcolor{bgcolor4}\color{blue} \bf 155.34 & \cellcolor{bgcolor4}\color{blue} \bf 93.29\\ \bottomrule
    \end{tabular}}
\end{table}

\begin{table}[!tb]
    \centering
    \caption{Perplexity scores on Wikitext-2 after training-free fine-tuning. The sparsity ratio is set to $60\%$ and $\alpha = 0.5$.}  
    \label{tab:ft_all}
    \resizebox{0.48\textwidth}{!}{
    \begin{tabular}{lcccccc}
    \toprule
    Base & FT & LlaMA2-7b & LlaMA2-13b & LlaMA3-8b\\ \midrule
    \algname{Dense} & - & 5.47 & 4.88 & 6.14\\ \hline
    \algname{Magnitude} & - & 6.9e3 & 10.10 & 4.05e5\\
    \algname{Magnitude} & \algname{DSnoT} & 4.1e3 & 10.19 & 4.18e4\\
    \rowcolor{bgcolor4}
    \algname{Magnitude} & \ft & \color{blue} \bf2.4e2 & \color{blue} \bf10.09 & \color{blue} \bf1.44e4\\ \midrule
    % \algname{Wanda} & - & 10.79 & 8.40 & 21.36\\
    \algname{Wanda} & - & \color{blue} \bf 9.72 & 7.75 & 21.36\\
    \algname{Wanda} & \algname{DSnoT} & 10.23 & \color{blue} \bf 7.69 & 20.70\\
    \rowcolor{bgcolor4}
    \algname{Wanda} & \ft & 10.08 & \color{blue} \bf7.69 & \color{blue} \bf20.50\\ \midrule
    \algname{RIA} & - & 10.29 & 7.85 & 21.09\\
    \algname{RIA} & \algname{DSnoT} & 9.97 & 7.82 & 19.51\\
    \rowcolor{bgcolor4}
    \algname{RIA} & \ft & \color{blue} \bf9.96 & \color{blue} \bf7.78 & \color{blue} \bf18.99\\ \bottomrule
    \end{tabular}}
\end{table}

\begin{table*}[!tb]
\centering
\caption{Accuracies (\%) for LLaMA2 models on 7 zero-shot tasks at 60\% unstructured sparsity.}
\label{tab:zsl_per_task_with_wandag_noDenseCompare}
\resizebox{0.8\textwidth}{!}{
\begin{tabular}{llcccccccc}
\toprule
Params & Method & BoolQ & RTE & HellaSwag & WinoGrande & ARC-e & ARC-c & OBQA & \cellcolor{bgcolor} Mean \\
\midrule
\multirow{7}{*}{LlaMA2-{7b}}
  %--- DENSE (unchanged) ---
 & Dense      
   & 77.7 & 62.8 & 57.2 & 69.2 & 76.4 & 43.4 & 31.4 & \cellcolor{bgcolor} 57.9 \\ \cmidrule{2-10}
 & \algname{Magnitude} & 41.2 & 51.3 & 37.0 & 55.7 & 50.0 & 27.0 & 16.2 & \cellcolor{bgcolor} 39.3\\
 & \algname{w. DSnoT} & 43.2 & \color{blue}54.2 & 38.4 & 56.4 & 53.3 & 27.7 & 20.6 & \cellcolor{bgcolor}  41.1 \\
 % \rowcolor{bgcolor4}
 &\cellcolor{bgcolor4}\bf \algname{w. $R^2$-DSnoT} & \cellcolor{bgcolor4}\color{blue}50.9 & \cellcolor{bgcolor4} 52.0 & \cellcolor{bgcolor4}\color{blue}39.8& \cellcolor{bgcolor4}\color{blue}56.8 & \cellcolor{bgcolor4}\color{blue}\color{blue}56.6 & \cellcolor{bgcolor4}\color{blue}28.3 & \cellcolor{bgcolor4}\color{blue}23.4 & \cellcolor{bgcolor}\color{blue} \bf43.4\\ \cmidrule{2-10}
 % & & 50.9 & 52.0 & 39.8 & 56.8 & 56.6 & 28.3 & 23.4 & 43.4 \\  
 % & \algname{Wanda} & 67.30& 53.10& 45.60& 64.80& 65.80& 32.80& 27.20& 50.90 \\
 % & \algname{w. DSnoT} & 65.20 & 53.40 & 44.10 & 63.90 & 65.00 & 31.40 & 26.00 & 49.90 \\
 % \rowcolor{bgcolor4}
 % &\cellcolor{bgcolor4}\bf \algname{w. $R^2$-DSnoT} & 65.20 & 53.10 & 44.20 & 64.00 & 65.60 & 31.80 & 25.60 & 49.90 \\ \cmidrule{2-10}
 & \algname{RIA} & \color{blue} 66.1 & 53.1 & 43.5 & 63.2 & 64.6 & 30.2 & 26.0 & \cellcolor{bgcolor} 49.5 \\
  & \algname{w. DSnoT} & 65.5 & 53.4 & \color{blue} 44.7 & 64.6 & \color{blue} 65.3 & \color{blue} 31.7 & 26.4 & \cellcolor{bgcolor} 50.2 \\
 % \rowcolor{bgcolor4}
 &\cellcolor{bgcolor4}\bf \algname{w. $R^2$-DSnoT} & \cellcolor{bgcolor4}65.2 & \cellcolor{bgcolor4}\color{blue} 53.8 & \cellcolor{bgcolor4}\color{blue} 44.7 & \cellcolor{bgcolor4}\color{blue} 65.1 & \cellcolor{bgcolor4}65.0 & \cellcolor{bgcolor4}31.6 & \cellcolor{bgcolor4}\color{blue} 27.0 & \cellcolor{bgcolor4}\cellcolor{bgcolor} \color{blue} \bf 50.3 \\ 
\hline
% \multirow{9}{*}{LlaMA2-{13b}}
%   %--- DENSE (unchanged) ---
%  & Dense      
%    &\\ \hline
%  & \algname{Wanda} & 76.91 & 56.32 & 50.53 & 67.88 & 70.75 & 38.23 & 28.80 & 55.63 \\
%  & \algname{w. DSnoT} & 76.73 & 55.96 & 49.88 & 67.48 & 70.79 & 35.58 & 27.40 & 54.83 \\
%  % \rowcolor{bgcolor4}
%  &\cellcolor{bgcolor4}\bf \algname{w. $R^2$-DSnoT}& 76.73 & 55.96 & 49.88 & 67.48 & 70.79 & 35.58 & 27.40 & 54.83 \\\cmidrule{2-10}
%  & \algname{RIA} & 77.40 & 58.48 & 50.02 & 68.11 & 70.54 & 36.77 & 28.80 & 55.73 \\
%   & \algname{w. DSnoT} & 76.54 & 59.57 & 49.22 & 67.40 & 70.20 & 36.01 & 27.80 & 55.25 \\
%  % \rowcolor{bgcolor4}
%  &\cellcolor{bgcolor4}\bf \algname{w. $R^2$-DSnoT} & 76.88 & 57.40 & 49.49 & 67.48 & 70.54 & 35.58 & 28.00 & 55.05 \\
%  \hline
 \multirow{7}{*}{LlaMA3-{8b}} 
  %--- DENSE (unchanged) ---
 & Dense      
   & 81.3 & 69.7 & 60.1 & 73.0 & 80.1 & 50.4 & 34.8 & \cellcolor{bgcolor} 64.2 \\ \cmidrule{2-10}
 & \algname{Magnitude} & 37.8 & 52.7 & 30.7 & 51.0 & 39.7 & 23.4 & 14.4 & \cellcolor{bgcolor} 35.7 \\
 & \algname{w. DSnoT} & 37.8 & 52.7 & \color{blue}33.4 & 49.9 & 43.5 & 23.0 & \color{blue}14.8 & \cellcolor{bgcolor} 36.4 \\
 % \rowcolor{bgcolor4}
 &\cellcolor{bgcolor4}\bf \algname{w. $R^2$-DSnoT} & \cellcolor{bgcolor4}37.8 & \cellcolor{bgcolor4}52.7 & \cellcolor{bgcolor4}33.1 & \cellcolor{bgcolor4}\color{blue}52.1 & \cellcolor{bgcolor4}\color{blue}43.9 & \cellcolor{bgcolor4}\color{blue}23.6 & \cellcolor{bgcolor4}\color{blue}14.8 & \cellcolor{bgcolor} \color{blue} \bf 37.1\\ \cmidrule{2-10}
 % && 37.8 & 52.7 & 33.1 & 52.1 & 43.9 & 23.6 & 14.8 & 37.1 \\  
 % & \algname{Wanda} & 70.31 & 53.43 & 39.51 & 62.04 & 59.85 & 28.92 & 21.60 & 47.95 \\
 % & \algname{w. DSnoT} & 69.79 & 53.43 & 39.52 & 60.85 & 61.66 & 28.24 & 21.60 & 47.87 \\
 % \rowcolor{bgcolor4}
 % &\cellcolor{bgcolor4}\bf \algname{w. $R^2$-DSnoT} & 69.54 & 53.43 & 39.74 & 61.33 & 60.65 & 28.67 & 20.60 & 47.71 \\ \cmidrule{2-10}
 & \algname{RIA} & 70.2 & 53.4 & 39.7 & 61.7 & 61.1 & \color{blue} 28.6 & 20.4 & \cellcolor{bgcolor} 47.9 \\
  & \algname{w. DSnoT} & \color{blue} 70.7 & 53.4 & \color{blue} 40.3 & 61.3 & \color{blue} 61.7 & 28.0 & 20.0 & \cellcolor{bgcolor} 47.9 \\
 % \rowcolor{bgcolor4}
 &\cellcolor{bgcolor4}\bf \algname{w. $R^2$-DSnoT} & \cellcolor{bgcolor4}70.4 & \cellcolor{bgcolor4}53.4 & \cellcolor{bgcolor4}\color{blue} \cellcolor{bgcolor4}40.3 & \cellcolor{bgcolor4}\color{blue} 61.9 & \cellcolor{bgcolor4}61.2 & \cellcolor{bgcolor4}28.3 & \cellcolor{bgcolor4}\color{blue} 21.0 & \cellcolor{bgcolor4} \cellcolor{bgcolor}\color{blue} \bf 48.1 \\
\bottomrule
\end{tabular}
} 
\end{table*}

% \begin{table}[!tb]
%     \centering
%     \resizebox{0.5\textwidth}{!}{
%     \begin{tabular}{l|ccccccccc}
%     Index & Base & FT & Reg. & LlaMA2-7b & LlaMA2-13b & LlaMA3-8b & OPT-1.3b\\ \hline
%     a) & Dense & \xmark & \xmark & 5.47 & 4.88 & 6.14 & 14.62\\ \hline
%     b) & \algname{Wanda} & \xmark & \xmark & 7.79 & 6.28 & 10.81 & 22.19\\
%     b1) & \algname{Wanda} & DSnoT & \xmark &\\
%     b2) (ours) & \algname{Wanda} & $R^2$-DSnoT & \xmark &\\
%     b3) (ours) & \algname{Wanda} & DSnoT & \cmark &\\
%     b4) (ours) & \algname{Wanda} & $R^2$-DSnoT & \cmark &\\ \hline
%     c) & \algname{RIA} & \xmark & \xmark & 6.81  & 5.93 & 9.34 & 18.08\\
%     c1) (ours) & \algname{RIA} & DSnoT & \xmark &\\
%     c2) (ours) & \algname{RIA} & $R^2$-DSnoT & \xmark &\\
%     c3) (ours) & \algname{RIA} & DSnoT & \cmark &\\
%     c4) (ours) & \algname{RIA} & $R^2$-DSnoT & \cmark &\\ \hline
%     \end{tabular}}
%     \caption{Perplexity scores on Wikitext-2 after training-free fine-tuning. The sparsity ratio is set to $0.5$ and $\alpha = 0.5$.}
%     \label{tab:ft_all}
% \end{table}

\paragraph{Various unstructured sparsity ratios.} 
We established a default unstructured sparsity ratio of 50\%. In this section, we investigate the impact of varying sparsity ratios, as detailed in \Cref{tab:various_sparsity}. For \algname{stochRIA}, we report the mean average perplexity after three trials. Given that \algname{stochRIA} has been shown to be stable, with variance examined in \Cref{tab:comparison}, we omit the variance to focus on performance. Our findings reveal that \algname{Wanda} is particularly sensitive to higher sparsity ratios, whereas both \algname{RIA} and our proposed \algname{stochRIA} demonstrate robustness to increased sparsity, maintaining stable performance across a broader range of conditions.
Interestingly, we observed that on LLaMA3-8b and OPT1.3b, \algname{stochRIA} consistently outperforms \algname{RIA}, whereas on LLaMA2-7b and LLaMA2-13b, the reverse is true. This intriguing phenomenon may be attributed to the heavy noise present in the sampling process for LLaMA3-8b and OPT1.3b. In such cases, selecting a subset of weights through \algname{stochRIA} may yield more reliable relative weight information, resulting in improved performance.

\subsection{Training-Free Fine-Tuning Comparisons} % This section examines various fine-tuning strategies, focusing on the impact of our novel approaches, specifically \emph{Relative and Regularized Dynamic Sparse no Training} (\ft) and a regularized decision boundary, detailed in (\ref{eqn:ft01}) and (\ref{eqn:ft02}). The perplexity results, presented in \Cref{tab:ft_all}, demonstrate significant performance enhancements through our strategies, which include relative reweighting and regularized boundaries, all without the need for additional training. Notably, improvements are especially significant when applied to non-fine-grained methods such as \algname{Magnitude}.

The intrinsic gap between pruned weights and the original, unpruned pretrained weights underscores the importance of minimizing reconstruction loss to achieve promising results. We introduced \ft, which incorporates relative weight reweighting and a regularized decision boundary during the dynamic sparse refinement step, all without additional training. Perplexity scores, as shown in \Cref{tab:ft_all}, reveal that our \ft approach consistently surpasses baseline methods and the previous state-of-the-art \algname{DSnoT} without fine-tuning. For instance, \algname{Magnitude} exhibited subpar perplexity scores on LlaMA2-7b and LlaMA3-8b; however, our \ft achieved perplexity reductions of 96.5\% and 96.4\%, respectively. These results not only validate \ft's efficacy but also offer guidance for scenarios involving high sparsity or underperforming pruned models, with minimal effort and no additional training.

\paragraph{Zero-shot performance.} To provide a comprehensive evaluation, we also conducted zero-shot classification tests using seven well-regarded datasets. These tests assess the pruned models' ability to accurately categorize objects or data points into previously unseen categories. We employed the methodology described by \cite{Wanda} and utilized tasks from the EleutherAI LM Harness \citep{gao2021framework}, including BoolQ \citep{clark2019boolq}, RTE \citep{wang2018glue}, HellaSwag \citep{zellers2019hellaswag}, WinoGrande \citep{sakaguchi2021winogrande}, ARC (Easy and Challenge) \citep{clark2018think}, and OpenbookQA \citep{mihaylov2018can}. The results, presented in \Cref{tab:zsl_per_task_with_wandag_noDenseCompare}, show that \ft consistently outperforms \algname{DSnoT} in zero-shot tasks, confirming its effectiveness. %To the best of our knowledge, \ft sets a new benchmark for training-free pruning and fine-tuning methods in terms of zero-shot performance.
To the best of our knowledge, \ft establishes a new state-of-the-art for training-free pruning and fine-tuning methods in zero-shot performance.

\section{Discussion and Future Work}
\textbf{Beyond pruning.} Our exploration of \algname{Wanda} and \algname{RIA} introduced the symmetric objective in (\ref{obj2}), initially aimed at post-training pruning for LLMs. However, our approach is extendable to post-training quantization and training-aware compression \citep{GPTQ, AQLM, PV-Tuning}, making these areas promising for future research.

\textbf{Better sampling.} In \Cref{sec:efficiency_stochastic_methods}, we demonstrated that selective sampling of matrix rows and columns enhances both performance and efficiency by maintaining diversity in lower-importance weights. Future research could explore asymmetric or non-uniform sampling within the (\ref{obj2}) framework to further optimize performance.

\textbf{Exploring symmetric designs.} As shown in \Cref{tab:comparison}, general and diagonal-specific symmetric designs for LLM compression highlight the potential of symmetric weight and activation patterns. Extending these approaches to distributed and federated settings \citep{kai2023fedp3, ye2024fedllm} could also be valuable.

\section{Conclusion}
\vspace{-2mm}
% This study systematically analyzed post-training pruning methods, particularly \algname{Wanda} and \algname{RIA}, showing that integrating input activations and weight importance significantly enhances pruning efficacy. We also introduced a training-free fine-tuning step based on relative weight importance within a prune-and-grow framework, surpassing existing baselines. These findings deepen our understanding of effective pruning strategies and lay the groundwork for future improvements in LLM compression.
% This study systematically analyzed post-training pruning methods, particularly \algname{Wanda} and \algname{RIA}, providing both empirical evidence and theoretical support for their efficacy. By introducing the symmetric objective in (\ref{obj2}), we highlighted the importance of input activations and weight significance in achieving more informed pruning decisions. Additionally, we proposed a training-free fine-tuning step based on relative weight importance within a prune-and-grow framework, surpassing existing baselines. These advancements not only enhance the theoretical understanding of post-training pruning but also lay the groundwork for future research on optimizing LLM compression through principled and efficient strategies.
This study systematically analyzed post-training pruning methods, particularly \algname{Wanda} and \algname{RIA}, providing both empirical evidence and theoretical insights into the role of input activations and relative weight importance through the symmetric objective in (\ref{obj2}). We also introduced a training-free fine-tuning step based on relative weight importance within a prune-and-grow framework, surpassing existing baselines. 
% These findings strengthen the theoretical foundation of post-training pruning and pave the way for further innovations in LLM compression.
% These advancements not only enhance the theoretical understanding of post-training pruning but also lay the groundwork for future research on optimizing LLM compression through principled and efficient strategies.
These advancements deepen the theoretical understanding of post-training pruning and pave the way for future research on efficient LLM compression strategies.

% \bibliography{example_paper}
% \bibliographystyle{icml2024}
% \newpage
% \clearpage
\bibliography{mybib}
\bibliographystyle{icml2024}

% \newpage
% \tableofcontents

%%%%%%%%%%%%%%%%%%%%%%%%%%%%%%%%%%%%%%%%%%%%%%%%%%%%%%%%%%%%%%%%%%%%%%%%%%%%%%%
%%%%%%%%%%%%%%%%%%%%%%%%%%%%%%%%%%%%%%%%%%%%%%%%%%%%%%%%%%%%%%%%%%%%%%%%%%%%%%%
% APPENDIX
%%%%%%%%%%%%%%%%%%%%%%%%%%%%%%%%%%%%%%%%%%%%%%%%%%%%%%%%%%%%%%%%%%%%%%%%%%%%%%%
%%%%%%%%%%%%%%%%%%%%%%%%%%%%%%%%%%%%%%%%%%%%%%%%%%%%%%%%%%%%%%%%%%%%%%%%%%%%%%%
\newpage
\appendix
\onecolumn
\section{Missing Proofs}
\subsection{Proof of \Cref{lemma:lm1}}
By using the definition of $g(\widetilde{\vW})$ in \Cref{obj1}, we have

$$
\begin{aligned}
g(\widetilde{\vW}) & =\sqrt{\sum_{k=1}^c\left\|\vX\left(\widetilde{\vW}_{: k}-\vW_{: k}\right)\right\|_2^2}+\sqrt{\sum_{j=1}^b\left\|\left(\widetilde{\vW}_{j:}-\vW_{j:}\right) \vY\right\|_2^2} \\
& =\sqrt{\sum_{k=1}^c \sum_{i=1}^a\left(\vX_{i:}\left(\widetilde{\vW}_{: k}-\vW_{: k}\right)\right)^2}+\sqrt{\sum_{j=1}^b \sum_{l=1}^d\left(\left(\widetilde{\vW}_{j:}-\vW_{j:}\right) \vY_{: l}\right)^2} \\
& =\sqrt{\sum_{k=1}^c \sum_{i=1}^a\left(\sum_{j=1}^b \vX_{i j}\left(\widetilde{\vW}_{j k}-\vW_{j k}\right)\right)^2}+\sqrt{\sum_{j=1}^b \sum_{l=1}^d\left(\sum_{k=1}^c\left(\widetilde{\vW}_{j k}-\vW_{j k}\right) \vY_{k l}\right)^2}
\end{aligned}
$$

Now say we want to prune away just a single weight $\vW_{j k}$. That is, we want to set $\widetilde{\vW}_{j k}=0$ and $\widetilde{\vW}_{j^{\prime} k^{\prime}}=\vW_{j^{\prime} k^{\prime}}$ for all $\left(j^{\prime}, k^{\prime}\right) \neq(j, k)$. For such a weight matrix $\widetilde{\vW}_{j k}$ the expression for $f(\widetilde{\vW})$ simplifies to

$$
\begin{aligned}
& g(\widetilde{\vW})=\sum_{i=1}^a\left(\sum_{j^{\prime}=1}^b \vX_{i j^{\prime}}\left(\widetilde{\vW}_{j^{\prime} k}-\vW_{j^{\prime} k}\right)\right)^2+\sum_{l=1}^d\left(\sum_{k^{\prime}=1}^c\left(\widetilde{\vW}_{j k^{\prime}}-\vW_{j k^{\prime}}\right) \vY_{k^{\prime} l}\right)^2 \\
& =\sqrt{\sum_{i=1}^a\left(\vX_{i j}\left(\widetilde{\vW}_{j k}-\vW_{j k}\right)+\sum_{j^{\prime} \neq j} \vX_{i j^{\prime}}\left(\widetilde{\vW}_{j^{\prime} k}-\vW_{j^{\prime} k}\right)\right)^2}\\
&\qquad +\sqrt{\sum_{l=1}^d\left(\left(\widetilde{\vW}_{j k}-\vW_{j k}\right) \vY_{k l}+\sum_{k^{\prime} \neq k}\left(\widetilde{\vW}_{j k}-\vW_{j k}\right) \vY_{k l}\right)^2} \\
& =\sqrt{\sum_{i=1}^a(\vX_{i j}\left(0-\vW_{j k}\right)+\sum_{j^{\prime} \neq j} \vX_{i j^{\prime}} \underbrace{\left(\vW_{j^{\prime} k}-\vW_{j^{\prime} k}\right)}_{=0})^2}\\
&\qquad +\sqrt{\sum_{l=1}^d(\left(0-\vW_{j k}\right) \vY_{k l}+\sum_{k^{\prime} \neq k} \underbrace{\left(\widetilde{\vW}_{j k}-\vW_{j k}\right)}_{=0} \vY_{k l})^2} \\
& =\sqrt{\sum_{i=1}^a\left(-\vX_{i j} \vW_{j k}\right)^2}+\sqrt{\sum_{l=1}^d\left(-\vW_{j k} \vY_{k l}\right)^2} \\
& =\sqrt{\sum_{i=1}^a \vX_{i j}^2 \vW_{j k}^2}+\sqrt{\sum_{l=1}^d \vW_{j k}^2 \vY_{k l}^2} \\
& =\left|\vW_{j k}\right|\left(\left\|\vX_{: j}\right\|_2+\left\|\vY_{k:}\right\|_2\right) \eqdef \vS_{j k}.
\end{aligned}
$$

\subsection{Proof of Theorem \ref{thm:main2}}
\begin{itemize}
    \item Assume it is possible to choose matrices $\vX \in \mbR^{a\times b}$ and $\vY \in \mbR^{c\times d}$ such that the identity 
            \begin{equation}\label{eqn1}
                \norm{\vX_{:k}}_2 + \norm{\vY_{j:}}_2 = \alpha_{jk} \eqdef \frac{1}{\norm{\vW_{j:}}_1}+ \frac{1}{\norm{\vW_{:k}}_1}
            \end{equation} 
            holds for all $j, k$. \emph{This is always possible!} 
            
            Indeed, if we choose $a=b$, and let the $j$-th row of $\vX$ be of the form $\vX_{:j} \eqdef t_{j} (1; \cdots; 1)\in \mbR^{b\times 1}$, where $t_j = \frac{1}{\sqrt{b} \norm{\vW_{j:}}_1}$, then $\norm{\vX_{j:}}_2 = t_j \sqrt{b} = \frac{1}{\norm{\vW_{j:}}_1}$. 
            
            Similarly, if we choose $d=c$, and let the $k$-th column of $\vY$ be of the form $\vY_{:k} \eqdef s_k(1,\cdots,1)\in \mbR^{1\times c}$, where $s_k = \frac{1}{\sqrt{c}\norm{\vW_{:k}}_1}$, then $\norm{\vY_{:k}}_2 = s_k\sqrt{c}= \frac{1}{\norm{\vW_{:k}}_1}$.
            
            So, \Cref{eqn1} holds. In this case, our score matrix \Cref{eqn0} reduces to the plug-and-play method \algname{RIA} \citep{RIA}.
\end{itemize}

\begin{itemize}
    \item Another (even simpler) possiblity for constructing matrices $\vX, \vY$ such that \Cref{eqn1} holds is as follows. Let $a=b$, and let $\vX = \Diag(\norm{\vW_{1:}}^{-1}_1, \cdots, \norm{\vW_{b:}}^{-1}_1)$. 
            Clearly, for all $j=1, \cdots, b$ we have $\norm{\vX_{j:}}_2 = \frac{1}{\norm{\vW_{j:}}_1}$. 
            
            Similarly, let $d=c$, and let $\vY = \Diag(\norm{\vW_{:1}}^{-1}_1, \cdots, \norm{\vW_{:c}}^{-1}_1)$. Clearly, for all $k=1, \cdots, c$, we have $\norm{\vY_{:k}}_2 = \frac{1}{\norm{\vW_{:k}}}_1$. 
            
            Therefore, $\norm{\vX_{:j}}_2 + \norm{\vY_{k:}}_2 = \frac{1}{\norm{\vW_{j:}}_1} + \frac{1}{\norm{\vW_{:k}}_1}$ for all $j, k$. So again, our score matrix (\ref{eqn0}) reduces to the plug-and-play method in \cite{RIA}.
\end{itemize}

\subsection{Proof of \Cref{lem:general2}}
Recall that in \Cref{sec:general_solution} $\vD_{\vX} \in \mbR^{b\times b}$ and $\vD_{\vY} \in \mbR^{c\times c}$ are diagonal matrices with entries defined as $\left(\mathbf{D}_{\mathbf{X}}\right)_{ii} = x_i = \left\|\mathbf{W}_{i:}\right\|_1^{-1}$ and $\left(\mathbf{D}_{\mathbf{Y}}\right)_{ii} = y_i = \left\|\mathbf{W}_{:i}\right\|_1^{-1}$ respectively, and $\mathbf{A}\in \mathbb{R}^{a\times b}$ and $\mathbf{B}\in \mbR^{c\times d}$ are arbitrary matrices. 
We first compute $\mathbf{A} \mathbf{D}_{\mathbf{X}}$. This product scales each column of $\mathbf{A}$ by the corresponding $x_i$. Specifically, for the $j$-th column, this operation is expressed as:
\[
\left(\mathbf{A} \mathbf{D}_{\mathbf{X}}\right)_{:j} = x_j \mathbf{A}_{:j}.
\]
The $\ell_2$-norm of this column is then given by:
\[
\left\|\left(\mathbf{A} \mathbf{D}_{\mathbf{X}}\right)_{:j}\right\|_2 = x_j \left\|\mathbf{A}_{:j}\right\|_2 = \frac{\left\|\mathbf{A}_{:j}\right\|_2}{\left\|\mathbf{W}_{j:}\right\|_1}.
\]

Next, we compute $\mathbf{D}_{\mathbf{Y}} \mathbf{B}$. In this computation, each row of $\mathbf{B}$ is scaled by the corresponding $y_i$. For the $k$-th row, the scaling is represented as:
\[
\left(\mathbf{D}_{\mathbf{Y}} \mathbf{B}\right)_{k:} = y_k \mathbf{B}_{k:}.
\]
The $\ell_2$-norm of this row is:
\[
\left\|\left(\mathbf{D}_{\mathbf{Y}} \mathbf{B}\right)_{k:}\right\|_2 = y_k \left\|\mathbf{B}_{k:}\right\|_2 = \frac{\left\|\mathbf{B}_{k:}\right\|_2}{\left\|\mathbf{W}_{:k}\right\|_1}.
\]

Finally, we consider the sum of these norms:
\[
\left\|\left(\mathbf{A} \mathbf{D}_{\mathbf{X}}\right)_{:j}\right\|_2 + \left\|\left(\mathbf{D}_{\mathbf{Y}} \mathbf{B}\right)_{k:}\right\|_2 = \frac{\left\|\mathbf{A}_{:j}\right\|_2}{\left\|\mathbf{W}_{j:}\right\|_1} + \frac{\left\|\mathbf{B}_{k:}\right\|_2}{\left\|\mathbf{W}_{:k}\right\|_1}.
\]

The first term involves scaling the $j$-th column of $\mathbf{A}$ by $x_j$, with the resulting norm being the original column norm divided by the $\ell_1$-norm of the corresponding weights in $\mathbf{W}$. Similarly, the second term scales the $k$-th row of $\mathbf{B}$ by $y_k$, with the resulting norm also being the original row norm divided by the $\ell_1$-norm of the corresponding weights in $\mathbf{W}$.

\subsection{Proof of \Cref{lem:generalized_p_norm}}
We aim to construct $\mathbf{X}_{: j}$ to be proportional to $\mathbf{W}_{j:}^{\top}$. A natural choice is to set
$$
\mathbf{X}_{: j} = c \cdot \mathbf{W}_{j:}^{\top},
$$
where $c$ is a scalar to be determined. A similar condition applies when considering $\mathbf{Y}_{k:}$. The central task is to compute the corresponding scaling factor $c$ for both $\mathbf{X}$ and $\mathbf{Y}$.

To determine $c$, we choose it such that
$$
\left\|\mathbf{X}_{: j}\right\|_2 = \left\|c \cdot \mathbf{W}_{j:}^{\top}\right\|_2 = \left\|\mathbf{W}_{j:}\right\|_p^{-1}.
$$

We now compute the $\ell_2$-norm of $\mathbf{X}_{: j}$:
$$
\left\|c \cdot \mathbf{W}_{j:}^{\top}\right\|_2 = |c| \cdot \left\|\mathbf{W}_{j:}^{\top}\right\|_2 = |c| \cdot \left\|\mathbf{W}_{j:}\right\|_2.
$$

Setting this equal to $\left\|\mathbf{W}_{j:}\right\|_p^{-1}$, we have:
$$
|c| \cdot \left\|\mathbf{W}_{j:}\right\|_2 = \left\|\mathbf{W}_{j:}\right\|_p^{-1}.
$$

Solving for $c$, we obtain:
$$
c = \frac{1}{\left\|\mathbf{W}_{j:}\right\|_p} \cdot \frac{1}{\left\|\mathbf{W}_{j:}\right\|_2}.
$$

Using this value of $c$, we define $\mathbf{X}_{: j}$ as:
$$
\mathbf{X}_{: j} = \frac{1}{\left\|\mathbf{W}_{j:}\right\|_p} \cdot \frac{1}{\left\|\mathbf{W}_{j:}\right\|_2} \cdot \mathbf{W}_{j:}^{\top}.
$$

This construction ensures that
$$
\left\|\mathbf{X}_{: j}\right\|_2 = \left\|\mathbf{W}_{j:}\right\|_p^{-1}.
$$

Similarly, for $\mathbf{Y}$, we have:
$$
\mathbf{Y}_{k:} = \frac{1}{\left\|\mathbf{W}_{:k}\right\|_p} \cdot \frac{1}{\left\|\mathbf{W}_{:k}\right\|_2} \cdot \mathbf{W}_{:k}^{\top},
$$
which satisfies \Cref{eqn:pnorm2}.

By combining these results, we conclude the proof of \Cref{lem:generalized_p_norm}.

\subsection{Proof of \Cref{lem:random_unit_vector_scaling}}

Let $\mathbf{u}$ be any unit vector in $\ell_2$-norm, i.e., $\norm{\mathbf{u}}_2 = 1$. Construct $\vX_{:j} = \norm{\vW_{j:}}_p^{-1} \mathbf{u}$. Then by using the definition of the $\ell_2$-norm, we have 

$$
\left\|\mathbf{X}_{: j}\right\|_2=\| \| \mathbf{W}_{j:}\left\|_p^{-1} \mathbf{u}\right\|_2=\left|\left\|\mathbf{W}_{j:}\right\|_p^{-1}\right|\|\mathbf{u}\|_2=\left\|\mathbf{W}_{j:}\right\|_p^{-1} \cdot 1=\left\|\mathbf{W}_{j:}\right\|_p^{-1}.
$$

Hence, we obtain $\left\|\mathbf{X}_{: j}\right\|_2=\left\|\mathbf{W}_{j:}\right\|_p^{-1}$, which is exactly as desired. 

Similarly, let $\mathbf{v}$ be any unit vector in $\ell_2$-norm, we have $|\vW_{jk}|\cdot \|\mathbf{W}_{:k}\|^{-1}_p$.

Put them together, we prove \Cref{lem:random_unit_vector_scaling}.

\subsection{Proof of \Cref{lem:stochria}}
Given that $\mathbf{X}_{:j}$ and $\mathbf{Y}_{k:}$ are vectors to be constructed, $\mathbf{W}$ is a matrix, and $S_j$ and $S_k$ are randomly sampled index sets from the $j$-th row and $k$-th column of $\mathbf{W}$, respectively, each with cardinality $\tau$, our task is to construct $\mathbf{X}_{:j}$ and $\mathbf{Y}_{k:}$ with specific norms. Specifically, the goal is to construct $\mathbf{X}_{:j}$ and $\mathbf{Y}_{k:}$ such that:
$$
\left\| \mathbf{X}_{:j} \right\|_2 + \left\| \mathbf{Y}_{k:} \right\|_2 = \frac{1}{\left\| \mathbf{W}_{j:S_j} \right\|_1} + \frac{1}{\left\| \mathbf{W}_{S_k:k} \right\|_1},
$$
where $\mathbf{W}_{j:S_j}$ denotes the entries of the $j$-th row of $\mathbf{W}$ at indices in $S_j$, and $\mathbf{W}_{S_k:k}$ denotes the entries of the $k$-th column of $\mathbf{W}$ at indices in $S_k$.

% We have the following construction steps.
% ### Step 1: Construct $\mathbf{X}_{:j}$

We first define the support vector $\mathbf{e}_{S_j}$ of appropriate size (equal to the number of rows in $\mathbf{X}$) as:
$$
(\mathbf{e}_{S_j})_i = \begin{cases}
    \frac{1}{\sqrt{\tau}}, & \text{if } i \in S_j, \\
    0, & \text{otherwise}.
\end{cases}
$$

The vector $\mathbf{e}_{S_j}$ has non-zero entries only at indices in $S_j$, each equal to $\frac{1}{\sqrt{\tau}}$, ensuring that the $\ell_2$-norm of $\mathbf{e}_{S_j}$ is 1:
$$
\left\| \mathbf{e}_{S_j} \right\|_2 = \sqrt{ \sum_{i \in S_j} \left( \frac{1}{\sqrt{\tau}} \right)^2 } = \sqrt{ \tau \cdot \left( \frac{1}{\sqrt{\tau}} \right)^2 } = 1.
$$

To construct $\mathbf{X}_{:j}$, we set:
$$
\mathbf{X}_{:j} = \frac{1}{\left\| \mathbf{W}_{j:S_j} \right\|_1} \cdot \mathbf{e}_{S_j}.
$$

A basic verification shows that the $\ell_2$-norm of $\mathbf{X}_{:j}$ is:
$$
\left\| \mathbf{X}_{:j} \right\|_2 = \frac{1}{\left\| \mathbf{W}_{j:S_j} \right\|_1} \cdot \left\| \mathbf{e}_{S_j} \right\|_2 = \frac{1}{\left\| \mathbf{W}_{j:S_j} \right\|_1} \cdot 1 = \frac{1}{\left\| \mathbf{W}_{j:S_j} \right\|_1}.
$$

% ### Step 2: Construct $\mathbf{Y}_{k:}$

Similarly, we define the support vector $\mathbf{e}_{S_k}$ of appropriate size (equal to the number of columns in $\mathbf{Y}$) as:
$$
(\mathbf{e}_{S_k})_i = \begin{cases}
    \frac{1}{\sqrt{\tau}}, & \text{if } i \in S_k, \\
    0, & \text{otherwise}.
\end{cases}
$$

To construct $\mathbf{Y}_{k:}$, we set:
$$
\mathbf{Y}_{k:} = \frac{1}{\left\| \mathbf{W}_{S_k:k} \right\|_1} \cdot \mathbf{e}_{S_k}^\top.
$$

% ### Step 3: Sum of Norms

Adding the norms:
$$
\left\| \mathbf{X}_{:j} \right\|_2 + \left\| \mathbf{Y}_{k:} \right\|_2 = \frac{1}{\left\| \mathbf{W}_{j:S_j} \right\|_1} + \frac{1}{\left\| \mathbf{W}_{S_k:k} \right\|_1},
$$
which matches the desired expression.

\textbf{Alternative construction using $\ell_1$ and $\ell_2$ norms.}

By definition:
$$
\left\| \mathbf{W}_{j:S_j} \right\|_1 = \sum_{i \in S_j} |w_{j i}|, \quad \left\| \mathbf{W}_{j:S_j} \right\|_2 = \sqrt{ \sum_{i \in S_j} w_{j i}^2 }.
$$

We can construct $\mathbf{X}_{:j}$ as:
$$
\mathbf{X}_{:j} = \frac{1}{\left\| \mathbf{W}_{j:S_j} \right\|_1} \cdot \frac{1}{\left\| \mathbf{W}_{j:S_j} \right\|_2} \cdot \mathbf{W}_{j:S_j}^\top,
$$
where $\mathbf{W}_{j:S_j}^\top$ is a vector with entries:
$$
(\mathbf{W}_{j:S_j}^\top)_i = \begin{cases}
    w_{j i}, & \text{if } i \in S_j, \\
    0, & \text{otherwise}.
\end{cases}
$$

Similarly, we can construct $\mathbf{Y}_{k:}$ as:
$$
\mathbf{Y}_{k:} = \frac{1}{\left\| \mathbf{W}_{S_k:k} \right\|_1} \cdot \frac{1}{\left\| \mathbf{W}_{S_k:k} \right\|_2} \cdot \mathbf{W}_{S_k:k}^\top,
$$
where $\mathbf{W}_{S_k:k}^\top$ is a vector with entries:
$$
(\mathbf{W}_{S_k:k}^\top)_i = \begin{cases}
    w_{i k}, & \text{if } i \in S_k, \\
    0, & \text{otherwise}.
\end{cases}
$$

Putting everything together, we prove \Cref{lem:stochria}.

% \textbf{Notes}

% \begin{itemize}
%     \item \textbf{Random Sampling:} $S_j$ and $S_k$ are randomly selected index sets from the $j$-th row and $k$-th column of $\mathbf{W}$, respectively. Each set has cardinality $\tau$.
%     \item \textbf{Normalization:} The vectors $\mathbf{e}_{S_j}$ and $\mathbf{e}_{S_k}$ are constructed to have $\ell_2$-norm equal to 1.
%     \item \textbf{Relation to $\mathbf{W}$:} The construction directly relates $\mathbf{X}_{:j}$ and $\mathbf{Y}_{k:}$ to the entries of $\mathbf{W}$ at the sampled indices.
% \end{itemize}

% Therefore, we have constructed $\mathbf{X}_{:j}$ and $\mathbf{Y}_{k:}$ such that the sum of their $\ell_2$-norms equals the sum of the inverses of the $\ell_1$-norms of the selected entries from $\mathbf{W}$.

\section{Symmetric Wanda Variant with Squared Frobenius Norms}\label{sec:squared_frobenius}
 
Choose $\varepsilon \in(0,1]$. Given $\vX \in \mathbb{R}^{a \times b}, \vW \in \mathbb{R}^{b \times c}$ and $\vY \in \mathbb{R}^{c \times d}$, define

$$
g^\prime(\widetilde{\vW}):=\|\vX(\widetilde{\vW}-\vW)\|_F^2+\|(\widetilde{\vW}-\vW) \vY\|_F^2,
$$

and consider solving the problem

$$
\begin{aligned}
\operatorname{mininimize} &\ \ g^\prime(\widetilde{\vW}) \quad s.t. & \operatorname{Mem}(\widetilde{\vW}) \leq \varepsilon \operatorname{Mem}(\vW), \widetilde{\vW} \in \mathbb{R}^{b \times c}.
\end{aligned}
$$

Note that

$$
\begin{aligned}
g^\prime(\widetilde{\vW}) & =\sum_{k=1}^c\left\|\vX\left(\widetilde{\vW}_{: k}-\vW_{: k}\right)\right\|_2^2+\sum_{j=1}^b\left\|\left(\widetilde{\vW}_{j:}-\vW_{j:}\right) \vY\right\|_2^2 \\
& =\sum_{k=1}^c \sum_{i=1}^a\left(\vX_{i:}\left(\widetilde{\vW}_{: k}-\vW_{: k}\right)\right)^2+\sum_{j=1}^b \sum_{l=1}^d\left(\left(\widetilde{\vW}_{j:}-\vW_{j:}\right) Y_{: l}\right)^2 \\
& =\sum_{k=1}^c \sum_{i=1}^a\left(\sum_{j=1}^b \vX_{i j}\left(\widetilde{\vW}_{j k}-\vW_{j k}\right)\right)^2+\sum_{j=1}^b \sum_{l=1}^d\left(\sum_{k=1}^c\left(\widetilde{\vW}_{j k}-\vW_{j k}\right) \vY_{k l}\right)^2
\end{aligned}
$$

Now say we want to prune away just a single weight $\vW_{j k}$. That is, we want to set $\widetilde{\vW}_{j k}=0$ and $\widetilde{\vW}_{j^{\prime} k^{\prime}}=\vW_{j^{\prime} k^{\prime}}$ for all $\left(j^{\prime}, k^{\prime}\right) \neq(j, k)$. For such a weight matrix $\widetilde{\vW}_{j k}$ the expression for $g^\prime(\widetilde{\vW})$ simplifies to

$$
\begin{aligned}
g^\prime(\widetilde{\vW}) & =\sum_{i=1}^a\left(\sum_{j^{\prime}=1}^b \vX_{i j^{\prime}}\left(\widetilde{\vW}_{j^{\prime} k}-\vW_{j^{\prime} k}\right)\right)^2+\sum_{l=1}^d\left(\sum_{k^{\prime}=1}^c\left(\widetilde{\vW}_{j k^{\prime}}-\vW_{j k^{\prime}}\right) \vY_{k^{\prime} l}\right)^2 \\
& =\sum_{i=1}^a\left(\vX_{i j}\left(\widetilde{\vW}_{j k}-\vW_{j k}\right)+\sum_{j^{\prime} \neq j} \vX_{i j^{\prime}}\left(\widetilde{\vW}_{j^{\prime} k}-\vW_{j^{\prime} k}\right)\right)^2\\
& \qquad +\sum_{l=1}^d\left(\left(\widetilde{\vW}_{j k}-\vW_{j k}\right) \vY_{k l}+\sum_{k^{\prime} \neq k}\left(\widetilde{\vW}_{j k}-\vW_{j k}\right) \vY_{k l}\right)^2 \\
& =\sum_{i=1}^a(\vX_{i j}\left(0-\vW_{j k}\right)+\sum_{j^{\prime} \neq j} \vX_{i j^{\prime}} \underbrace{\left(\vW_{j^{\prime} k}-\vW_{j^{\prime} k}\right)}_{=0})^2+\sum_{l=1}^d(\left(0-\vW_{j k}\right) \vY_{k l}+\sum_{k^{\prime} \neq k} \underbrace{\left(\widetilde{\vW}_{j k}-\vW_{j k}\right)}_{=0} \vY_{k l})^2 \\
& =\sum_{i=1}^a\left(-\vX_{i j} \vW_{j k}\right)^2+\sum_{l=1}^d\left(-\vW_{j k} \vY_{k l}\right)^2 \\
& =\sum_{i=1}^a \vX_{i j}^2 \vW_{j k}^2+\sum_{l=1}^d \vW_{j k}^2 \vY_{k l}^2 \\
& =\vW_{j k}^2\left(\left\|\vX_{: j}\right\|_2^2+\left\|Y_{k:}\right\|_2^2\right) \eqdef \vS_{j k}^2.
\end{aligned}
$$

Our proposal is to choose entry $(j, k)$ which the smallest score $\vS_{j k}$. Special cases:

1. If we choose $\vX=\mathbf{0} \in \mathbb{R}^{a \times b}$, then our pruning method reduces to "output" \algname{Wanda}:

$$
\vS_{j k}:=\left|\vW_{j k}\right|\left\|\vY_{k:}\right\|_2
$$

2. If we choose $\vY=\mathbf{0} \in \mathbb{R}^{c \times d}$, then our pruning method reduces to "input" \algname{Wanda}:

$$
\vS_{j k}:=\left|\vW_{j k}\right|\left\|\vX_{: j}\right\|_2.
$$

3. If we choose $\vX=\vW^{\top} \in \mathbb{R}^{c \times b}(a=c)$ and $\vY=\vW^{\top} \in \mathbb{R}^{c \times b}(d=b)$, then our score matrix becomes

$$
\vS_{j k} \stackrel{(27)}{=}\left|\vW_{j k}\right| \sqrt{\left\|\vX_{: j}\right\|_2^2+\left\|\vY_{k:}\right\|_2^2}=\left|\vW_{j k}\right| \sqrt{\left\|\vW_{j:}\right\|_2^2+\left\|\vW_{: k}\right\|_2^2}
$$

Letting $\vG_{j k}^2:=\frac{1}{b+c}\left(\left\|\vW_{j:}\right\|_2^2+\left\|\vW_{: k}\right\|_2^2\right)$, note that

$$
\begin{aligned}
\|\vG\|_F^2 & =\sum_{j=1}^b \sum_{k=1}^c \vG_{j k}^2 \\
& =\frac{1}{b+c} \sum_{j=1}^b \sum_{k=1}^c\left(\left\|\vW_{j:}\right\|_2^2+\left\|\vW_{: k}\right\|_2^2\right) \\
& =\frac{1}{b+c}\left(\sum_{j=1}^b \sum_{k=1}^c\left\|\vW_{j:}\right\|_2^2+\sum_{k=1}^c \sum_{j=1}^b\left\|\vW_{: k}\right\|_2^2\right) \\
& =\frac{1}{b+c}\left(c \sum_{j=1}^b\left\|\vW_{j:}\right\|_2^2+b \sum_{k=1}^c\left\|\vW_{: k}\right\|_2^2\right) \\
& =\frac{1}{b+c}\left(c\|\vW\|_F^2+b\|\vW\|_F^2\right) \\
& =\|\vW\|_F^2
\end{aligned}
$$

Clearly,

$$
\frac{\vS_{j k}^2}{(b+c)\|\vW\|_F^2} \stackrel{}{=} \frac{\vW_{j k}^2 \vG_{j k}^2}{\|\vW\|_F^2}
$$

4. Assume it is possible to choose matrices $\vX \in \mathbb{R}^{a \times b}$ and $\vY \in \mathbb{R}^{c \times d}$ such that the identity

$$
\sqrt{\left\|\vX_{j:}\right\|_2^2+\left\|\vY_{: k}\right\|_2^2}=\alpha_{j k}:=\frac{1}{\left\|\vW_{j:}\right\|_1}+\frac{1}{\left\|\vW_{: k}\right\|_1}
$$

holds for all $j, k$ (note that this is not always possible!). In this case, our score matrix reduces to the plug-and-play method of \cite{RIA}.

\section{Additional Experiments}
\subsection{Implementation Details}
Our selected baselines are implemented using the source code from \algname{Wanda}\footnote{\url{https://github.com/locuslab/wanda/tree/main}} and \algname{RIA}\footnote{\url{https://github.com/biomedical-cybernetics/Relative-importance-and-activation-pruning}}. The default settings remain unchanged to ensure consistency. Notably, we explicitly set the sequence length to 2048 instead of using the maximum possible length to enable a fair comparison, following the strategy outlined in \algname{RIA}.

The training-free fine-tuning component is based on \algname{DSnoT}\footnote{\url{https://github.com/zyxxmu/DSnoT}}. We configure the maximum cycle count to 50 and set the update threshold to 0.1. The default power of variance for regrowing and pruning is set to 1. Additionally, we incorporate the regularized relative design, resulting in our modified approach, \algname{DSnoT}.

The seed for sampling the calibration data is set to 0. For N:M structural pruning, to enable an intuitive comparison, we use the standard approach without employing channel reallocation or linear sum assignment, as used in \algname{RIA}.

\subsection{Optimal $\ell_p$ Norm}\label{sec:optimal_p}
In this study, we further explore the influence of the $\ell_p$ norm, considering standard norms where $p \in [1, 2, 3, 4]$, as well as the $0$-norm and $\infty$-norm. The results are presented in \Cref{tab:p_norms}. We observed that higher $p$ values degrade performance, as reflected by the perplexity scores, with $p=1$ yielding the best results. This may be due to the fact that in pruning, significantly magnifying the differences between weights is not beneficial. Additionally, we found that both the $0$-norm and $\infty$-norm do not yield promising results, as they capture only partial, and often highly biased, information about the weights.

\begin{table}[!tb]
    \centering
    \caption{\parbox{0.45\textwidth}{Perplexity scores on Wikitext-2 for \algname{p-norm}. The sparsity ratio is $50\%$, and all results correspond to $\alpha=1$. }}
    \label{tab:p_norms}
    \resizebox{0.5\textwidth}{!}{
    \begin{tabular}{l|ccccccc}
    \toprule
    p & LlaMA2-7b & LlaMA2-13b & LlaMA3-8b & OPT-1.3b\\ \midrule
    1 & \textbf{6.88} & \textbf{5.95} & \textbf{9.44} & \textbf{18.95}\\
    2 & 6.90 & 5.96 & 9.48 & 19.02\\
    3 & 6.95 & 6.01 & 9.57 & 19.66\\
    4 & 7.12 & 6.08 & 9.92 & 20.77\\ \midrule
    0 & 7.78 & 6.28 & 10.81 & 22.17\\
    $\infty$ & 8.60 & 6.80 & 11.28 & 24.92\\ \bottomrule
    \end{tabular}}
\end{table}

\subsection{$\ell_p$ Norm Re-weighting}\label{sec:norm_p_reweighting}
In this section, we explore different $\ell_p$ norm re-weighting strategies. Our default re-weighting approach is defined in \Cref{eqn:pnorm2} and is referred to as $\mathrm{S1}$. Additionally, we investigate alternative strategies, denoted as $\mathrm{S2}$, $\mathrm{S3}$, and $\mathrm{S4}$, as specified below:

% \vspace{-5mm}
\begin{equation}
\begin{aligned}
    \mathrm{S2} \eqdef \mathbf{S}_{jk} &= |\mathbf{W}_{jk}| / (\|\mathbf{W}_{j:}\|_p + \|\mathbf{W}_{:k}\|_p),\\
    \mathrm{S3} \eqdef \mathbf{S}_{jk} &= |\mathbf{W}_{jk}| \cdot (\|\mathbf{W}_{j:}\|_p + \|\mathbf{W}_{:k}\|_p),\\
    \mathrm{S4} \eqdef \mathbf{S}_{jk} &= |\mathbf{W}_{jk}| / (\|\mathbf{W}_{j:}\|^{-1}_p + \|\mathbf{W}_{:k}\|^{-1}_p).\\ \notag
\end{aligned}
\end{equation}
% \vspace{-10mm}

The comparative results for these strategies are presented in \Cref{tab:p_norm_reweighting}. As shown, our default strategy ($\mathrm{S1}$) achieves the best performance, while the alternative designs fail to deliver improvements.

\begin{table}[!tb]
    \centering
    \caption{\parbox{0.55\textwidth}{Perplexity scores on Wikitext-2 for \algname{$\ell_p$-norm} re-weighting with different strategies. The sparsity ratio is $50\%$, and all results are computed with $\alpha=0.5$ and $p=1$.}}
    \label{tab:p_norm_reweighting}
    \resizebox{0.6\textwidth}{!}{
    \begin{tabular}{l|cccc}
    \toprule
    Strategy & LLaMA2-7b & LLaMA2-13b & LLaMA3-8b & OPT-1.3b\\ \midrule
    $\mathrm{S1}$ (default) & 6.81 & 5.83 & 9.34 & 18.08 \\
    $\mathrm{S2}$ & 6.99 & 5.91 & 9.58 & 19.01 \\
    $\mathrm{S3}$ & 9.32 & 6.87 & 17.31 & 31.66 \\
    $\mathrm{S4}$ & 14.51 & 20.78 & 30.47 & 53.17 \\ \bottomrule
    \end{tabular}}
\end{table}

We hypothesize that the performance differences arise due to the relative magnitudes of the terms $\|\mathbf{W}_{j:}\|_p + \|\mathbf{W}_{:k}\|_p$ and $\|\mathbf{W}_{j:}\|^{-1}_p + \|\mathbf{W}_{:k}\|^{-1}_p$. Specifically, we assume that $\|\mathbf{W}_{j:}\|_p + \|\mathbf{W}_{:k}\|_p$ is typically large, while $\|\mathbf{W}_{j:}\|^{-1}_p + \|\mathbf{W}_{:k}\|^{-1}_p$ is generally small. Consequently, dividing by the former ($\mathrm{S2}$) or multiplying by the latter ($\mathrm{S4}$) reduces the magnitude of the pruning weights. We will provide statistical evidence to validate this assumption in subsequent sections.

\subsection{Influence of Sampling Ratios}\label{sec:sampling_ratios}
In this section, we examine the impact of varying sampling ratios in \algname{stochRIA}. It is important to note that these ratios are applied over $\min(b, c)$, where $b$ and $c$ represent the number of rows and columns in each layer, respectively. In \Cref{tab:stoch_res_diff_sampling_ratio}, we can see the performance of \algname{stochRIA} is generally stable and compares favorably to that of \algname{RIA} when sampling across entire rows and columns, particularly for $\beta \geq 0.05$. At this threshold and above, the performance is robust, occasionally even surpassing less noisy sampling configurations. However, at an extremely low ratio of $\beta = 0.01$, there is a significant performance decline. Consequently, we have set $\beta = 0.1$ as the default setting for our experiments.

\begin{table}[!tb]
    \centering
    \caption{\parbox{0.5\textwidth}{Perplexity scores on Wikitext-2 for \algname{stochRIA} with different sampling ratios. The sparsity ratio is $50\%$, and all results correspond to $\alpha=1$. We highlight those performance drops over 0.1 as significant.}}
    \label{tab:stoch_res_diff_sampling_ratio}
    \resizebox{0.55\textwidth}{!}{
    \begin{tabular}{l|ccccccc}
    \toprule
    ratio ($\beta$) & LlaMA2-7b & LlaMA2-13b & LlaMA3-8b & OPT-1.3b\\ \midrule 
    1 & 6.91 & 5.95 & 9.45 & 18.88\\     \midrule
    0.9 & 6.91 & 5.95 & 9.43 & 18.87\\
    0.5 & 6.90 & 5.95 & 9.42 & 18.84\\
    0.1 & 6.91 & 5.95 & 9.46 & 18.78\\
    0.05 & 6.91 & 5.96 & 9.47 & 18.91\\
    0.01 & 6.98 & 6.00 & 9.69 {\small \color{red} -0.24} & 19.36 {\small \color{red} -0.48}\\
    \bottomrule
    \end{tabular}}
\end{table}

\begin{table}[!tb]
    \centering
    \caption{\parbox{0.65\textwidth}{\ft Hyperparameter Ablations on LLaMA3-8b. Each row shows the non-default hyperparameter values compared to the best-performing method.}}
    \label{tab:ft_hyper_abl}
    \resizebox{0.7\textwidth}{!}{
    \begin{tabular}{l|c|cccccc}
    \toprule
    base & setting & $p$ & grow relative? & $\gamma_1$ & prune relative? & $\gamma_2$ & perplexity$\downarrow$\\ \midrule
    \multirow{8}{*}{\algname{Wanda}} & best & 2 & \cmark & 0 & \xmark & 0.0001 & 18.99\\ \cmidrule{2-8}
      & \multirow{2}{*}{$p$} & 1 & & & & & 19.04\\ 
      & & $\infty$ & &&&& 18.99\\ \cmidrule{2-8}
      & \multirow{2}{*}{$\gamma$} & & & & & 0 & 18.99\\ 
      & & & &&& 0.001 & 18.99\\ \cmidrule{2-8}
      & \multirow{3}{*}{relative} & & \xmark & & \xmark && 19.49\\
      & & & \xmark & & \cmark && 19.25\\
      & & & \cmark & & \cmark && 19.63\\ \cmidrule{1-8}
    \multirow{8}{*}{\algname{RIA}} & best & 2 & \xmark & 0 & \cmark & 0.001 & 20.50\\ \cmidrule{2-8}
      & \multirow{2}{*}{$p$} & 1 & & & & & 25.61\\ 
      & & $\infty$ & &&&& 20.51\\ \cmidrule{2-8}
      & \multirow{2}{*}{$\gamma$} & & & & & 0 & 20.51\\ 
      & & & &&& 0.0001 & 20.52\\ \cmidrule{2-8}
      & \multirow{3}{*}{relative} & & \xmark & & \xmark && 21.33\\
      & & & \cmark & & \xmark && 22.16\\
      & & & \cmark & & \cmark && 22.60\\
      \bottomrule
    \end{tabular}}
\end{table}
\subsection{Analysis of \ft Hyperparameters}\label{sec:ablation_ft_parameters}
In \Cref{sec:training_free_fine_tuning}, we introduced the equations for our proposed \ft method, specifically \Cref{eqn:ft01} and \Cref{eqn:ft02}. This method primarily involves three key hyperparameters: the regularization penalty $\gamma_1, \gamma_2$ and the norm type $p$. Additionally, we consider whether to apply relative importance reweighting during the growing or pruning phases—or during both. Given the number of hyperparameters, understanding their interactions can be computationally expensive and time-consuming.

To address this complexity, we adopt a systematic approach by performing a random search over 20 different combinations of hyperparameter settings. These combinations include: $p \in \{1, 2, \infty\}$, $\gamma_1 \in \{0, 0.0001, 0.001\}$, $\gamma_2 \in \{0, 0.0001, 0.001\}$, and binary choices for relative reweighting (True/False) during both the growing and pruning phases. For each of the 20 trials on the same model, we identify the best-performing combination and treat its hyperparameters as the "ground truth." We then evaluate the behavior under different scenarios and report the results in \Cref{tab:ft_hyper_abl}.

Our findings reveal several notable insights:

\begin{itemize}
    \item Norm type $p$: The smooth $\ell_p$-norm with $p = 2$ consistently achieves the best performance. Compared to the non-differentiable $\ell_1$-norm, which underperforms due to its non-smooth nature, and the $\ell_{\infty}$-norm, which focuses only on the largest values and ignores smaller differences, the $\ell_p$-norm with $p = 2$ balances sensitivity and robustness effectively.
    
    \item  Relative importance reweighting: Applying relative reweighting during either the growing or pruning phase improves performance significantly—yielding a 0.5 improvement on \algname{Wanda} and 0.83 on \algname{RIA}. However, applying reweighting to both phases simultaneously leads to substantial performance degradation, with a 0.64 and 2.1 drop on \algname{Wanda} and \algname{RIA}, respectively.
    
    \item  Regularization penalty $\gamma$: The impact of $\gamma$ is minimal, as variations in its value result in only marginal differences in performance. This finding highlights the greater importance of the relative reweighting strategy.
\end{itemize}

% \kai{TODO: R2-DSnoT is undefined.}

\end{document}